\documentclass[letterpaper]{article} 
\usepackage{aaai2026}  
\usepackage{times}  
\usepackage{helvet}  
\usepackage{courier}  
\usepackage[hyphens]{url}  
\usepackage{graphicx} 
\usepackage{natbib}  
\usepackage{caption} 
\usepackage{algorithm}
\usepackage{algorithmic}

\usepackage{newfloat}
\usepackage{listings}
\DeclareCaptionStyle{ruled}{labelfont=normalfont,labelsep=colon,strut=off} 
\floatstyle{ruled}
\newfloat{listing}{tb}{lst}{}
\floatname{listing}{Listing}
\usepackage{multirow}
\usepackage{amssymb} 
\usepackage{amsmath}
\usepackage{booktabs}

\def\ie{\emph{i.e., }}

\def\name{CyC3D}
\def\fig{Fig. }

\title{\name{}: Fine-grained Controllable 3D Generation via Cycle Consistency Regularization}

\author{
Hongbin Xu\textsuperscript{\rm 1}, Chaohui Yu\textsuperscript{\rm 2}, Feng Xiao\textsuperscript{\rm 1}, Jiazheng Xing\textsuperscript{\rm 3}, Hai Ci\textsuperscript{\rm 3}, Weitao Chen\textsuperscript{\rm 4}, Fan Wang\textsuperscript{\rm 2}, Ming Li\textsuperscript{\rm 5}
}
\affiliations{
\textsuperscript{\rm 1}South China University of Technology
\textsuperscript{\rm 2}Alibaba Group
\textsuperscript{\rm 3}National University of Singapore
\textsuperscript{\rm 4}Fudan University
\textsuperscript{\rm 5}Guangdong Laboratory of Artificial Intelligence and Digital Economy (SZ)
}
\usepackage{bibentry}

\begin{document}

\maketitle

\begin{abstract}
Despite the remarkable progress of 3D generation, achieving controllability, i.e., ensuring consistency between generated 3D content and input conditions like edge and depth, remains a significant challenge. 
Existing approaches often struggle to maintain accurate alignment, leading to noticeable discrepancies.
To address this issue, we propose \name{}, a new framework designed to enhance controllable 3D generation by explicitly encouraging cyclic consistency during training between the second-order 3D content, generated based on extracted signals from the first-order generation, and its original input controls.
Specifically, we employ an efficient feed-forward backbone that can generate a 3D object from an input condition and a text prompt.
Given an initial viewpoint and a control signal, a novel view is rendered from the generated 3D content, from which the extracted condition is used to regenerate the 3D content.
This re-generated output is then rendered back to the initial viewpoint, followed by another round of control signal extraction, forming a cyclic process with two consistency constraints.
\emph{View consistency} ensures coherence between the two generated 3D objects, measured by semantic similarity to accommodate generative diversity. 
\emph{Condition consistency} aligns the final extracted signal with the original input control, preserving structural or geometric details throughout the process.
Extensive experiments on zero-shot GSO/ABO benchmarks demonstrate that \name{} significantly improves controllability, especially for fine-grained details, outperforming existing methods across various conditions (e.g., +14.17\% PSNR for edge, +6.26\% PSNR for sketch).
\end{abstract}


\section{Introduction}



\begin{figure*}[t]
\centering
\includegraphics[width=\linewidth]{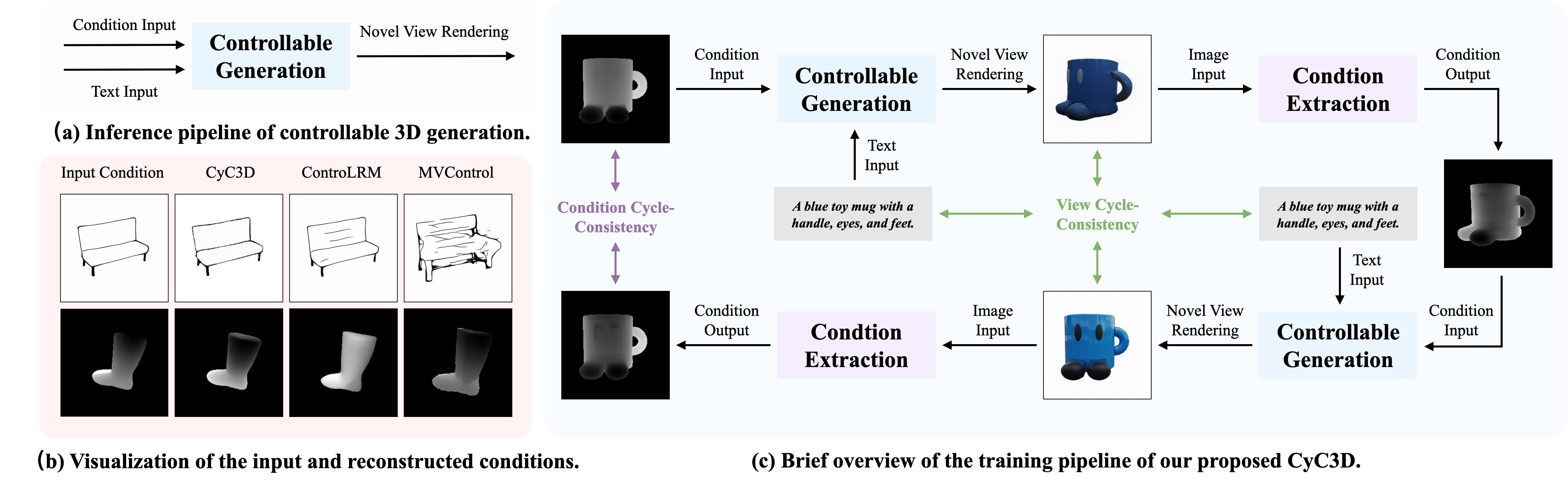}
\vspace{-0.6cm}
\caption{Brief overview of our proposed \textbf{CyC3D}.}
\vspace{-0.6cm}
\label{fig:motivation}
\end{figure*}

The creation of 3D models from text descriptions (text-to-3D) or image collections (image-to-3D) is a fundamental task in computer graphics, attracting growing interest from a wide range of domains, including VR/AR \cite{behravan2025generative}, digital game design \cite{xu2024sketch2scene}, and filmmaking \cite{wang2025cinemaster}. 
The mainstream approaches can be broadly categorized into two groups: optimization-based methods and feed-forward methods. Optimization-based 3D generation methods leverage Score Distillation Sampling (SDS) \cite{poole2022dreamfusion} to iteratively refine 3D objects based on textual or visual inputs, effectively distilling the rich prior knowledge embedded in large pre-trained diffusion models \cite{rombach2022high}. In contrast, feed-forward 3D generation approaches \cite{nichol2022point,tang2023volumediffusion,hong20243dtopia,hong2023lrm,zou2024triplane,tochilkin2024triposr,chen2025lara} generate 3D content in a single inference pass, significantly improving efficiency and making real-time applications more feasible.

While significant progress has been made in 3D generation research, a critical yet under-explored challenge remains—\emph{the controllability of 3D generation}. Controllable generation seeks to integrate precise control signals, such as edges, sketches, depth maps, surface normals, and text prompts, as conditioning inputs. By incorporating these structured constraints, controllable 3D generation enhances the flexibility and reliability of the synthesis process, enabling more precise manipulation of the generated content.

To advance research in controllable 3D generation, MVControl \cite{li2024controllable} extends ControlNet, originally designed for conditioned 2D image generation, to a multi-view diffusion model \cite{shi2023mvdream}. It first predicts four-view images, which are then processed by a multi-view Gaussian reconstruction model \cite{tang2025lgm} to generate the final 3D representation. In contrast, ControLRM \cite{xu2024controlrm} aims to develop an end-to-end controllable 3D generation framework by replacing the original image encoder of a large reconstruction model (LRM) \cite{hong2023lrm} with a conditional encoder, enabling more precise conditioning. 
The inference pipeline of these controllable 3D generation works is summarized in \fig \ref{fig:motivation} (a).
While these studies have demonstrated a degree of controllability, a significant challenge remains, \ie, achieving precise and fine-grained conditional control. 
As illustrated in \fig \ref{fig:motivation} (b), the extracted signals from the 3D models generated by MVControl and ControLRM exhibit noticeable discrepancies from the original input controls, underscoring the need for more refined and accurate controllable 3D generation.

To address this challenge, we propose \name{}, a novel and general training framework that enhances the controllability of 3D generation. Our approach extends a feed-forward 3D generation network into a cyclic paradigm, enabling more precise and consistent adherence to input control signals, as illustrated in \fig \ref{fig:motivation} (c).
Specifically, given an input control condition on a reference view, the generated 3D content is first rendered from randomly sampled new viewpoints. From these rendered images, the corresponding control conditions are extracted and fed back into the generation model as inputs, ensuring that the model learns to maintain control consistency across views. Subsequently, the model regenerates the 3D content, which is then rendered back to the initial reference viewpoint, where the extracted control conditions should match the original input condition. To further refine controllability, we introduce two complementary consistency constraints: \textit{View Cycle Consistency}, which ensures that the two generated contents have similar semantics in embedded space since they are generated with the same prompt and highly coupled conditions, and \textit{Condition Cycle Consistency}, which enforces strict alignment between the extracted control signals from the final rendered images and the original input condition.
We compare our \name{} against state-of-the-art approaches on GSO and ABO datasets \cite{xu2024controlrm} for zero-shot evaluation. Experimental results demonstrate that our framework significantly enhances controllability across all benchmarks while preserving competitive generation quality compared to existing methods.

\begin{figure*}[t]
\centering
\includegraphics[width=0.9\linewidth]{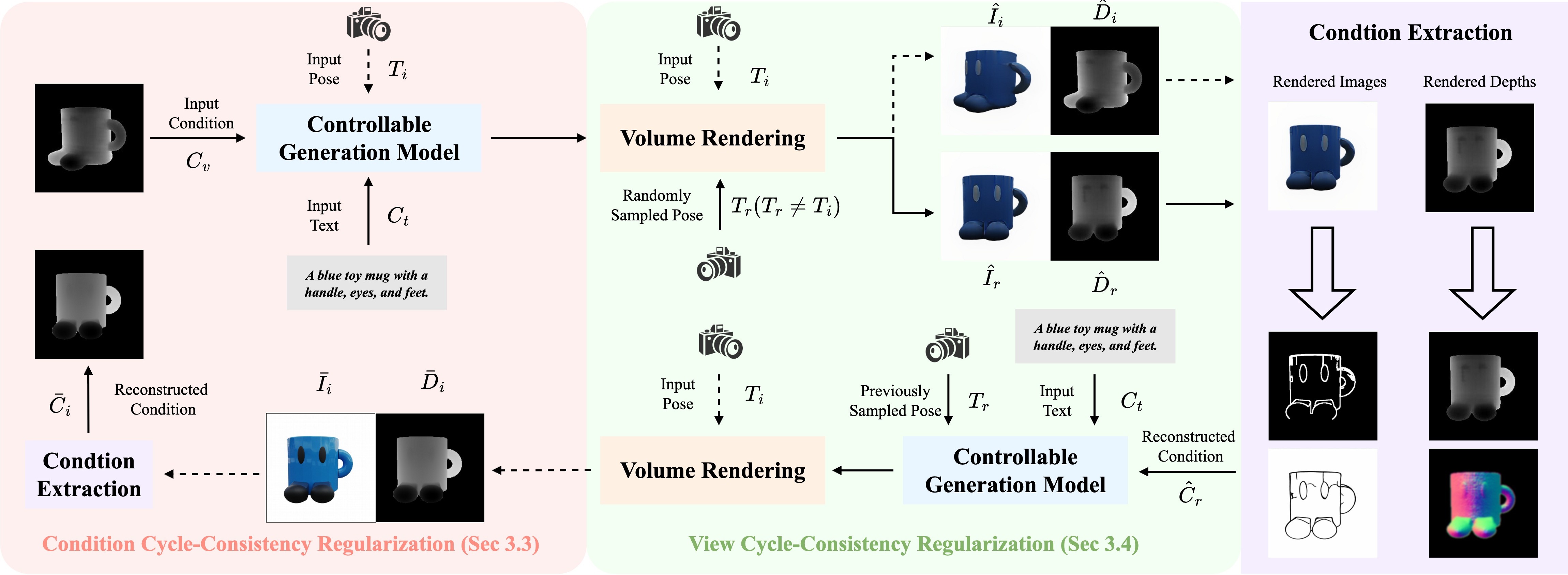}
\vspace{-0.3cm}
\caption{Training framework of our proposed \name{}. It contains two kinds of regularization: condition cycle consistency regularization (Sec. \ref{sec:method:condition_cycle_consistency}) and view cycle consistency regularization (Sec. \ref{sec:method:view_cycle_consistency}).}
	\vspace{-0.6cm}
	\label{fig:method}
\end{figure*}

We highlight the contributions of this work as follows:
(1) \emph{New Insight and Framework:} We reveal that existing efforts in controllable 3D generation still perform poorly on fine-grained controllability, and propose a novel framework (\name{}) to handle the problem, which extends the original single-stage novel view synthesis pipeline to a double-stage one with cycle consistency.
(2) \emph{Cyclic Consistency Feedback:} To refine controllability, we propose two complementary consistency constraints: condition cycle consistency and view cycle consistency. The former regularizes the consistency between input and reconstructed conditions, the latter enhances the semantic consistency of different stages.
(3) \emph{Evaluation and Promising Results:} We provide a comprehensive zero-shot evaluation of the controllability on GSO/ABO benchmarks, and demonstrate that \name{} comprehensively outperforms existing methods.

\section{Related Work}

\noindent\textbf{Text-/Image-to-3D Generation.}
3D generation can be broadly categorized into optimization-based and feed-forward approaches:
(1) Optimization-based methods offer a viable alternative to traditional techniques by circumventing the need for extensive text-to-3D datasets. DreamFusion \cite{poole2022dreamfusion} exemplifies this approach by utilizing SDS loss to optimize neural fields with diffusion priors, resulting in high-quality 3D assets. 
Despite their effectiveness, these methods often encounter the Janus problem, which arises from inconsistencies between 2D supervision and 3D geometry.
(2) Feed-forward models have emerged as a prominent approach for efficient 3D generation, leveraging extensive 3D datasets. Techniques such as LRM \cite{hong2023lrm} and TripoSR \cite{tochilkin2024triposr} utilize transformer architectures to predict NeRF from single-view images, demonstrating high generalization capabilities. Multi-view methods extend these approaches by synthesizing higher-quality 3D models from single-view images using multi-view diffusion models, as demonstrated by SyncDreamer \cite{liu2023syncdreamer} and MVDream \cite{shi2023mvdream}. 
Though Cycle3D \cite{tang2025cycle3d} builds a cycle-consistent pipeline of diffusion and reconstruction models, the whole pipeline is built in the inference model. Whereas our cycle-consistency is used for training a real-time feed-forward generation model.

\noindent\textbf{Controllable 3D Generation.}
Achieving controllability in 3D generation remains a critical objective in the field. Techniques developed for controllable 2D generation have provided valuable insights and methods that are applicable to the more complex domain of controllable 3D generation. Inspired by ControlNet \cite{zhang2023adding}, MVControl \cite{li2024controllable} integrates a trainable control network with a multi-view diffusion model to facilitate controllable multi-view image generation. This method combines coarse Gaussian generation with SDS-based refinement but faces challenges related to model generalization and procedural complexity. ControLRM \cite{xu2024controlrm} addresses these limitations with its end-to-end feed-forward model, ensuring rapid inference and enhanced control precision, thus offering an efficient and effective solution for controllable 3D generation. Building on the advances of ControlNet++ \cite{li2025controlnet} and ControLRM \cite{zhang2023adding}, we improve conditional controls to achieve more sophisticated and nuanced control over the generated 3D assets.

\section{Method}


\subsection{Generation Backbone}
\label{sec:method:backbone}

For controllable 3D generation, our \name{} can be combined with arbitrary feed-forward backbones (e.g. ControLRM \cite{xu2024controlrm}, MVControl \cite{li2024controllable}).
In default, we select ControLRM \cite{xu2024controlrm} as our backbone model.
ControLRM contains a 2D condition extractor and a 3D triplane decoder.
The conditional generator in ControLRM is designed to take both text and 2D visual conditions as inputs. This generator can utilize either a transformer or diffusion backbone to create initial 2D latent representations, which are subsequently transformed into 3D models.
The 3D triplane decoder is inherited from a pre-trained large reconstruction model \cite{hong2023lrm}.
The triplane transformer decoder employs a unique representation of 3D data by decomposing it into three orthogonal planes (X, Y, and Z). Each plane is processed using transformer layers, which utilize self-attention mechanisms to capture spatial relationships and dependencies within the data.

\begin{figure*}[t]
	\centering
	\includegraphics[width=\linewidth]{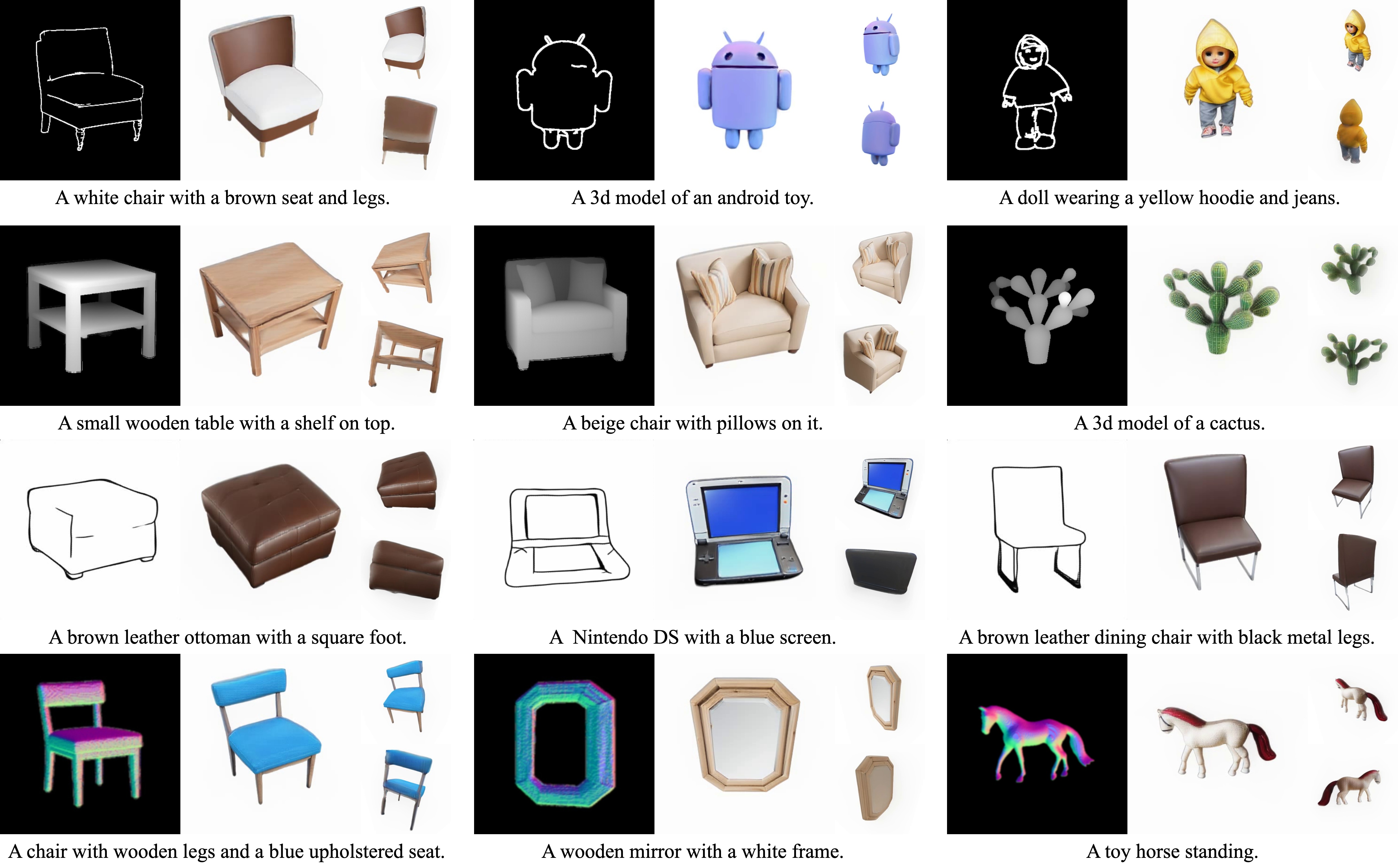}
	\vspace{-0.6cm}
	\caption{Visualization results of our \name{}. The input conditions and the rendered images at different views are visualized. Our \name{} can generate high-quality results with consistent conditional controls.}
	\vspace{-0.6cm}
	\label{fig:visualization}
\end{figure*}

\subsection{Condition Extraction}
\label{sec:method:condition_extractor}

To ensure the gradient back-propagation of cycle consistency in our CyC3D framework during training, the utilized condition extractor should be fully differentiable.
The design of the condition extractor for each condition control is shown as follows:

\noindent\textbf{Canny Condition.}
We extract edge maps using the Canny algorithm \cite{canny1986computational}.
Following \cite{li2025controlnet}, a differentiable Canny extractor can be implemented based on Kornia \cite{riba2020kornia} to extract edge maps.

\noindent\textbf{Sketch Condition.}
Following ControlNet \cite{zhang2023adding}, we utilize a pre-trained CNN network to extract sketch conditions from images.
Since the CNN network is differentiable, the gradients can be passed backward through the network.

\noindent\textbf{Depth Condition.}
In analogy with the sketch condition, we also utilize a pre-trained foundation model (Depth Anything \cite{yang2024depth}) to extract the depth condition from rendered images.
As an alternative, we can also normalize the rendered depth map via volume rendering to extract the geometric prior from the 3D content.

\noindent\textbf{Normal Condition.}
In practice, we find that the existing normal estimation models \cite{bae2021estimating} utilized in ControlNet \cite{zhang2023adding} tend to predict over-smoothed normal maps.
It may result in a large gap between the exact 3D shape of the rendered multi-view images and the estimated normal maps.
Consequently, we adopt an alternative solution to predict the depth map first and then convert the depth map to a normal map via a differentiable module \cite{huang2021m3vsnet}.
As an alternative, we also extract the normal from the rendered depth maps via volume rendering.

\subsection{Condition Cycle Consistency Regularization}
\label{sec:method:condition_cycle_consistency}

Denote the input edge condition as $C_e \in \mathbb{R}^{H \times W}$, the sketch condition as $C_s \in \mathbb{R}^{H \times W}$, the depth condition as $C_d \in \mathbb{R}^{H \times W}$, and the normal condition as $C_n \in \mathbb{R}^{H \times W \times 3}$.
The input visual condition is noted as $C_v \in \{C_e, C_s, C_d, C_n\}$.
Given the text prompt $C_t$ and the input visual condition as $C_v$, the generation model outputs the triplane representation $P \in \mathbb{R}^{3 \times C_p \times H_p \times W_p}$ corresponding to the input conditions.
\begin{equation}
\setlength{\abovedisplayskip}{0pt}
\setlength{\belowdisplayskip}{0pt}
\small
	P_i = f_{\text{gen}} (C_v, C_t, T_i),
        \label{eq1}
\end{equation}
where $T_i$ is the camera pose of the input condition, which is usually assumed to be an identity matrix after normalizing all the camera extrinsics to a frontal view.
$f_{\text{gen}}$ is the backbone generation model for controllable 3D generation.

Given a 3D point $x \in \mathbb{R}^3$, the color and density filed $(c, \sigma)$ can be obtained via sampling the corresponding features on the triplane $P_i$: $(c, \sigma)  = f_{\text{mlp}} (P_i, x)$.
By sampling the points along the viewpoint $T_r \in \mathbb{R}^{4 \times 4}$, we can render the image on view $T_r$ via volume rendering \cite{mildenhall2021nerf}:
\begin{equation}
\small
	\hat{I_r}, \hat{D_r} = f_{\text{render}} (P_i, T_r),
    \label{eq2}
\end{equation}
where $T_r$ is the randomly sampled viewpoints as shown in Fig. \ref{fig:method}. $\hat{I_r}$ is the rendered image on view $T_r$, and $\hat{D_r}$ is the rendered depth map.
Then we can use the corresponding condition extractor $f_{\text{cond}}$ introduced in Sec. \ref{sec:method:condition_extractor} and Fig. \ref{fig:method} (b) to extract the condition map on the sampled view $T_r$:
\begin{equation}
\small
	\hat{C_r} = f_{\text{cond}} (\hat{I_r}),
    \label{eq3}
\end{equation}
where $\hat{C_r}$ is the extracted condition map. 
Note that the condition extractor $f_{\text{cond}}$ is differentiable.

In a second time, we can feed the obtained condition map $\hat{C_r}$ on novel view $T_r$ back to the generation model $f_{\text{gen}}$ again:
\begin{equation}
\small
	P_r = f_{\text{gen}} (\hat{C_r}, C_t, T_r),
    \label{eq4}
\end{equation}
where the predicted triplane $P_r$ can further be used to render on the input reference view $T_i$:
\begin{equation}
\small
	\overline{I_i}, \overline{D_i} = f_{\text{render}} (P_r, T_i),
    \label{eq5}
\end{equation}
where $\overline{I_i}$ is the rendered image on the input reference view, and $\overline{D_i}$ is the rendered depth map.
Then we can extract the condition map from the rendered image $\overline{I_i}$:
\begin{equation}
\small
	\overline{C_i} = f_{\text{cond}} (\overline{I_i}),
	\label{eq6}
\end{equation}
where $\overline{C_i}$ is the generated condition map back on the input reference view.

As shown in \fig \ref{fig:method}, we can calculate the difference between the reconstructed condition and the input condition to regularize the fine-grained controls.
Depending on the type of conditional controls, we also propose two different kinds of conditional cycle-consistency feedback as follows:

\noindent\textbf{2D Conditional Control Feedback.}
If we only consider 2D conditional controls extracted from the rendered images (e.g., Canny map extracted from RGB image), we can use a condition extractor to reconstruct the control signal.
The generated condition map $\overline{C_i}$ via Eq. \ref{eq6} should be consistent with the original condition map $C_v$ on the input reference view.
\begin{equation}
\small
	L_{\text{cond}} = \sum_{T_r} \| \overline{C_i} - C_v \|_2^2.
	\label{eq7}
\end{equation}
If the randomly sampled camera pose $T_r$ equals the input viewpoint $T_i$, the two-step inference in cycle consistency will be degraded to a single one (as shown in Fig. \ref{fig:method} (c)), saving the computation cost:
\begin{equation}
\small
	\hat{I_i}, \hat{D_i} = f_{\text{render}} (f_{\text{gen}} (C_v, C_t, T_i), T_i),
	\label{eq8}
\end{equation}
\begin{equation}
\small
	\hat{C_i} = f_{\text{cond}} (\hat{I_i}),
    \label{eq9}
\end{equation}
where $\hat{C_i}$ is the extracted condition map on the input reference view.
In this way, we can modify Eq. \ref{eq7} as:
\begin{equation}
\small
	L_{\text{cond}} = \sum_{T_r \neq T_i} \| \overline{C_i} - C_v \|_2^2 + \| \hat{C_i} - C_v \|_2^2.
	\label{eq10}
\end{equation}

\begin{figure*}[t]
	\centering
	\includegraphics[width=0.9\linewidth]{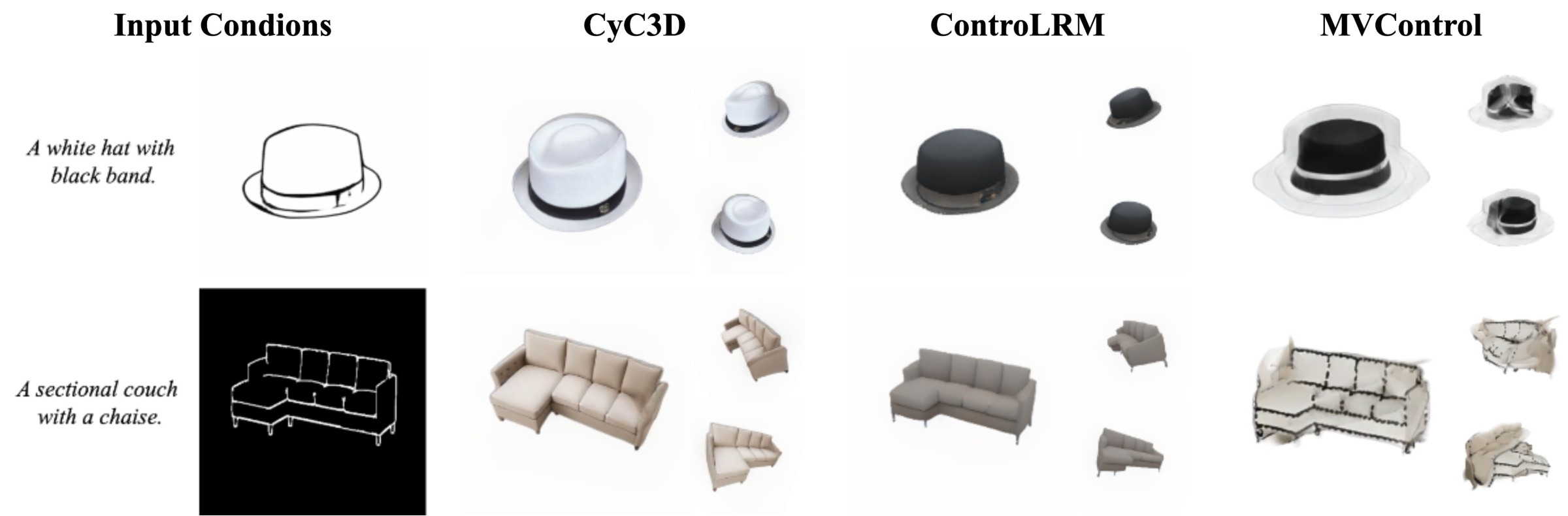}
	\vspace{-0.3cm}
	\caption{Qualitative comparison among our CyC3D and the other state-of-the-art controllable 3D generation methods. In the 1st row, the baseline methods fail to generate consistent content with the text prompts. In the 2nd row, ControLRM generates a fainter color compared with our \name{}.}
	\vspace{-0.6cm}
	\label{fig:qualitative}
\end{figure*}

\noindent\textbf{3D Conditional Control Feedback.}
If we consider 3D conditional controls, such as depth or normal conditions, the 3D shape of the generated objects should also be regularized for fine-grained control.

(1) If the input condition is depth map ($C_v = C_d$), the explicit consistency on the rendered depth map and the input conditional depth map can be calculated as follows:
\begin{equation}
\small
	L_{\text{cond-d}} = \sum_{T_r} \| f_{\text{norm}} (\overline{D_i}) - C_d \|_2^2,
	\label{eq11}
\end{equation}
where $f_{\text{norm}}$ is the depth normalizing function.
Since the rendered depth $\overline{D_i}$ and the input depth $C_d$ have a gap in the scale of depth, we utilize $f_{\text{norm}}$ to normalize them to the same scale and calculate a scale-agnostic loss following \cite{ranftl2020towards}.

In analogy with Eq. \ref{eq10}, when the randomly sampled camera pose $T_r$ equals the input viewpoint $T_i$, there is no need to run the inference twice during propagation.
We can modify Eq. \ref{eq11} as follows:
\begin{equation}
\small
	L_{\text{cond-d}} = \sum_{T_r \neq T_i} \| f_{\text{norm}} (\overline{D_i}) - C_d \|_2^2 + \|  f_{\text{norm}} (\hat{D_i}) - C_d \|_2^2,
	\label{eq12}
\end{equation}
where $\hat{D_i}$ can be obtained from Eq. \ref{eq8}.

(2) If the input condition is a normal map ($C_v = C_n$), the explicit depth-normal consistency can be further calculated as follows:
\begin{equation}
\small
	L_{\text{cond-n}} = \sum_{T_r} \| f_{\text{d2n}} (\overline{D_i}) - C_n \|_2^2,
	\label{eq13}
\end{equation}
where $f_{\text{d2n}}$ is a differentiable module \cite{huang2021m3vsnet} that can convert the depth map to a unified normal map.

In analogy with Eq. \ref{eq10} and \ref{eq12}, when the randomly sampled camera pose $T_r$ equals the input viewpoint $T_i$, there is no need to run the inference twice during propagation.
We can thus modify Eq. \ref{eq13} as follows:
\begin{equation}
\small
	L_{\text{cond-n}} = \sum_{T_r \neq T_i} \| f_{\text{d2n}} (\overline{D_i}) - C_n \|_2^2 + \| f_{\text{d2n}} (\hat{D_i}) - C_n \|_2^2,
    \label{eq14}
\end{equation}
where $\hat{D_i}$ can be obtained from Eq. \ref{eq8}.

\begin{table*}[t]
	\centering
	\resizebox{0.9\linewidth}{!}{
		\begin{tabular}{l|ccc|ccc|ccc|cc}
			\toprule
			\multirow{2}{*}{\textbf{Methods}} & \multicolumn{3}{c|}{\textbf{Edge }}          & \multicolumn{3}{c|}{\textbf{Sketch}}                & \multicolumn{3}{c|}{\textbf{Depth}}               & \multicolumn{2}{c}{\textbf{Normal}}       \\
			& \textbf{PSNR$^\uparrow$} & \textbf{SSIM$^\uparrow$} & \textbf{MSE$^\downarrow$} & \textbf{PSNR$^\uparrow$} & \textbf{SSIM$^\uparrow$} & \textbf{MSE$^\downarrow$} & \textbf{M-MSE$^\downarrow$} & \textbf{Z-MSE$^\downarrow$} & \textbf{R-MSE$^\downarrow$} & \textbf{NB-MSE$^\downarrow$} & \textbf{DN-CON$^\downarrow$} \\ \midrule \midrule
			\textbf{GSGEN \cite{chen2024text}}                    & 13.01           & 0.780           & 0.0526          & 15.07           & 0.772           & 0.0322          & 0.1277          & 0.1268          & 0.0311          & 0.0230          & 0.0497                  \\
			\textbf{GaussianDreamer \cite{yi2024gaussiandreamer}}          & 12.41           & 0.766           & 0.0597          & 14.58           & 0.792           & 0.0362          & 0.0740          & 0.0947          & 0.0391          & 0.0225          & 0.0366                  \\
			\textbf{DreamGaussian \cite{tang2023dreamgaussian}}           & 9.94            & 0.686           & 0.1048          & 14.42           & 0.765           & 0.0380          & 0.0980          & 0.0916          & 0.0406          & 0.0216          & 0.0335                  \\
			\textbf{VolumeDiffusion \cite{tang2023volumediffusion}}          & 13.25           & 0.818           & 0.0497          & 16.50           & 0.836           & 0.0231          & 0.1202          & 0.0953          & 0.0531          & 0.0190          & 0.0172                  \\
			\textbf{3DTopia \cite{hong20243dtopia}}                  & 9.63            & 0.704           & 0.1134          & 15.54           & 0.801           & 0.0287          & 0.1193          & 0.1064          & 0.0420          & 0.0268          & 0.0372                  \\
			\textbf{MVControl \cite{li2024controllable}}                & 10.44           & 0.723           & 0.0952          & 14.51           & 0.746           & 0.0370          & 0.0555          & 0.0645          & 0.0160          & 0.0209          & 0.0315                  \\
			\textbf{ControLRM-T \cite{xu2024controlrm}}              & 12.43           & 0.836           & 0.0607          & 17.17           & 0.836           & 0.0202          & 0.0567          & 0.0713          & 0.0599          & 0.0120          & 0.0359                  \\
			\textbf{ControLRM-D \cite{xu2024controlrm}}              & 12.47           & 0.837           & 0.0600          & 17.23           & 0.837           & 0.0197          & 0.0383          & 0.0546          & 0.0650          & 0.0119          & 0.0357                  \\ \midrule
			\textbf{CyC3D-T (Ours)}              & \textbf{15.17}  & \textbf{0.885}  & \textbf{0.0342} & 18.31  & 0.889  & 0.0159 & \textbf{0.0051} & \textbf{0.0037} & 0.0065          & \textbf{0.0027} & 0.0062                  \\
			\textbf{CyC3D-D (Ours)}              & 15.09           & 0.885           & 0.0347          & \textbf{18.38}  & \textbf{0.890}  & \textbf{0.0156} & \textbf{0.0051} & 0.0041          & \textbf{0.0060} & 0.0028          & \textbf{0.0060}          \\ \bottomrule
            
		\end{tabular}
	}
	\vspace{-0.2cm}
	\caption{Quantitative comparison of controllability with state-of-the-art controllable 3D generation methods on \textbf{GSO} benchmark. $\uparrow$ denotes higher result is better, while $\downarrow$ means lower is better. The best results are highlighted in \textbf{bold}. We provide the results of four different conditions: edge, sketch, depth, and normal.}
	\vspace{-0.4cm}
	\label{tab:controllability:gso}
\end{table*}

\begin{table*}[t]
	\centering
	\resizebox{0.9\linewidth}{!}{
		\begin{tabular}{l|ccc|ccc|ccc|cc}
			\toprule
			\multirow{2}{*}{\textbf{Methods}} & \multicolumn{3}{c|}{\textbf{Edge }}          & \multicolumn{3}{c|}{\textbf{Sketch}}                & \multicolumn{3}{c|}{\textbf{Depth}}               & \multicolumn{2}{c}{\textbf{Normal}}       \\
			& \textbf{PSNR$^\uparrow$} & \textbf{SSIM$^\uparrow$} & \textbf{MSE$^\downarrow$} & \textbf{PSNR$^\uparrow$} & \textbf{SSIM$^\uparrow$} & \textbf{MSE$^\downarrow$} & \textbf{M-MSE$^\downarrow$} & \textbf{Z-MSE$^\downarrow$} & \textbf{R-MSE$^\downarrow$} & \textbf{NB-MSE$^\downarrow$} & \textbf{DN-CON$^\downarrow$} \\ \midrule \midrule
			\textbf{GSGEN \cite{chen2024text}}                    & 12.75           & 0.737           & 0.0614          & 14.25           & 0.744           & 0.0384          & 0.1004          & 0.0967          & 0.0463          & 0.0171          & 0.0479                  \\
			\textbf{GaussianDreamer \cite{yi2024gaussiandreamer}}          & 11.92           & 0.736           & 0.0699          & 13.75           & 0.734           & 0.0439          & 0.0809          & 0.1061          & 0.0521          & 0.0188          & 0.0463                  \\
			\textbf{DreamGaussian \cite{tang2023dreamgaussian}}           & 9.38            & 0.683           & 0.1178          & 12.90           & 0.716           & 0.0523          & 0.1267          & 0.0762          & 0.0433          & 0.0188          & 0.0349                  \\
			\textbf{VolumeDiffusion \cite{tang2023volumediffusion}}          & 12.58           & 0.734           & 0.0619          & 15.03           & 0.780           & 0.0321          & 0.1007          & 0.1198          & 0.0671          & 0.0188          & 0.0264                  \\
			\textbf{3DTopia \cite{hong20243dtopia}}                  & 8.48            & 0.633           & 0.1469          & 14.41           & 0.741           & 0.0372          & 0.1916          & 0.1844          & 0.0559          & 0.0346          & 0.0418                  \\
			\textbf{MVControl \cite{li2024controllable}}                & 9.52            & 0.656           & 0.1161          & 13.34           & 0.686           & 0.0479          & 0.0454          & 0.0467          & 0.0181          & 0.0133          & 0.0350                  \\
			\textbf{ControLRM-T \cite{xu2024controlrm}}              & 10.49           & 0.552           & 0.0909          & 15.77           & 0.777           & 0.0272          & 0.0514          & 0.0497          & 0.0700          & 0.0136          & 0.0572                  \\
			\textbf{ControLRM-D \cite{xu2024controlrm}}              & 11.51           & 0.592           & 0.0725          & 15.80           & 0.778           & 0.0269          & 0.0493          & 0.0545          & 0.0661          & 0.0158          & 0.0548                  \\ \midrule
			\textbf{CyC3D-T (Ours)}              & 12.98           & \textbf{0.756}  & 0.0540          & \textbf{16.55}  & \textbf{0.847}  & \textbf{0.0225} & 0.0080          & \textbf{0.0078} & 0.0102          & \textbf{0.0029} & \textbf{0.0102}         \\
			\textbf{CyC3D-D (Ours)}              & \textbf{12.99}  & 0.755           & \textbf{0.0539} & 16.45           & 0.845           & 0.0231          & \textbf{0.0075} & 0.0087          & \textbf{0.0091} & 0.0030          & 0.0109                      \\ \bottomrule
		\end{tabular}
	}
	\vspace{-0.2cm}
	\caption{Quantitative comparison of controllability with state-of-the-art controllable 3D generation methods on \textbf{ABO} benchmark. $\uparrow$ denotes higher result is better, while $\downarrow$ means lower is better. The best results are highlighted in \textbf{bold}. We provide the results of four different conditions: edge, sketch, depth, and normal.}
	\vspace{-0.4cm}
	\label{tab:controllability:abo}
\end{table*}

\begin{figure}[t]
\centering
\includegraphics[width=\linewidth]{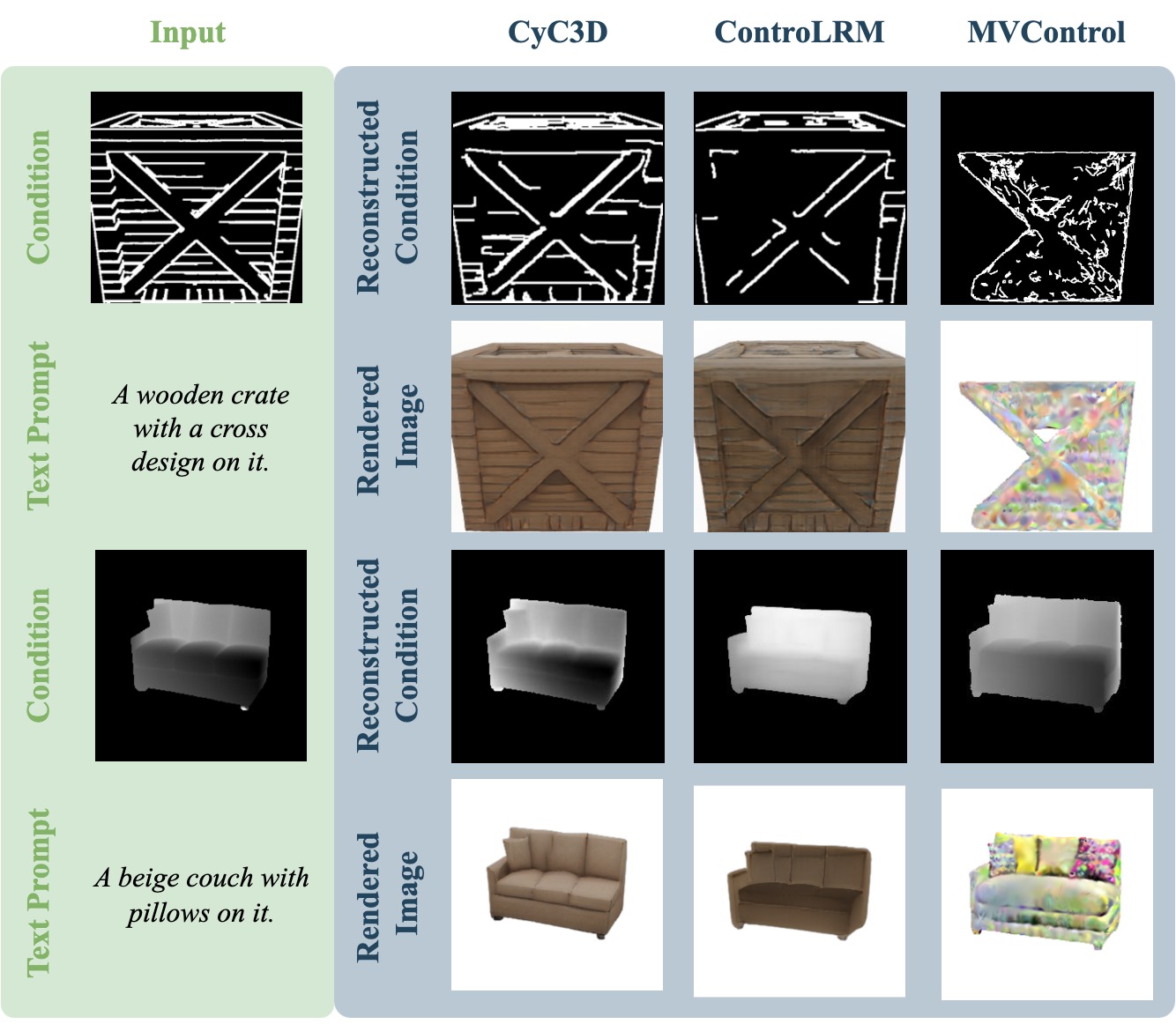}
\vspace{-0.6cm}
\caption{Qualitative comparison of controllability among our \name{} and the existing state-of-the-art methods in controllable 3D generation.}
\vspace{-0.6cm}
\label{fig:qualitative_controllability}
\end{figure}

\subsection{View Cycle Consistency Regularization}
\label{sec:method:view_cycle_consistency}

Since the condition cycle consistency regularization (Sec. \ref{sec:method:condition_cycle_consistency}) is a strong constraint that enforces the reconstructed and the input conditions to be as similar as possible, only using this loss might lead to overfitting towards the input viewpoint.
If the generated 3D content from the initial view fails, the remaining processes might be misled to a wrong cross-view mapping.
Only using the condition cycle consistency on the input view can not handle this issue, and the semantic multi-view consistency might be ignored during optimization.
Consequently, we propose View Cycle Consistency Regularization to ensure convergence.

\noindent\textbf{Rendering Regularization Loss.}
To regularize the possible failures in generation, the image reconstruction loss between the rendered images from generated 3D contents and the ground truth images on different views is used to supervise the generation model $f_{\text{gen}}$:
\begin{equation}
\small
	L_{\text{render}} = \sum_{T_l} \| f_{\text{render}}(f_{\text{gen}}(C_v,C_i,T_i), T_l) - I_l^{\text{gt}}\|_2^2,
    \label{eq15}
\end{equation}
where $T_l$ means the pre-defined ground truth viewpoints in the multi-view dataset, and $I_l^{\text{gt}}$ is the corresponding ground truth image.
Note that this constraint can only be used in the first-time feedforward computation of the generation model.
Because the randomness of the generation model might be depressed if enforcing the first-time and second-time generated 3D contents to be the same.

\noindent\textbf{Semantic Consistency Loss}:
We utilize the text prompt $C_t$ to regularize the semantic consistency of different stages.
\begin{equation}
\begin{aligned}
\small
    L_{\text{clip}} = \sum_{T_r} (1 - \cos (f_{\text{clip-t}} (C_t), f_{\text{clip-i}} (\hat{I}_r))) + \\ (1 - \cos (f_{\text{clip-t}} (C_t), f_{\text{clip-i}} (\overline{I}_i))).
    \label{eq16}
\end{aligned}
\end{equation}

\noindent\textbf{View Consistency Regularization:}
The overall view cycle consistency loss can be computed as follows:
\begin{equation}
\small
    L_{\text{view}} = L_{\text{render}} + \alpha \cdot L_{\text{clip}},
    \label{eq17}
\end{equation}
where $\alpha$ is the set to 5.0 following \cite{xu2024controlrm} to balance the scale of different terms as default.

\subsection{Overall Loss}
\label{sec:method:overall_loss}

Finally, the overall loss is the combination of the aforementioned losses:
\begin{equation}
	L_{\text{total}} = 
	\left\{
	\begin{aligned}
		& L_{\text{view}} + \lambda \cdot L_{\text{cond}} + \beta \cdot L_{\text{cond-d}}, \;\; \text{if} \;\;  C_v = C_d  \\
		& L_{\text{view}} + \lambda \cdot L_{\text{cond}} + \beta \cdot L_{\text{cond-n}}, \;\; \text{if} \;\; C_v = C_n \\
		& L_{\text{view}} + \lambda \cdot L_{\text{cond}}, \quad \text{otherwise}
	\end{aligned}
	\right.
\end{equation}
where $\lambda$ is set to 1.0 in default, and $\beta$ is set to 0.1 for regularization.

\section{Experiment}

\subsection{Implementation Details}
\label{sec:experiment:implementation}


\noindent\textbf{Datasets.}
Our training dataset is the training split of the G-Objaverse (\textbf{GOBJ}) dataset \cite{qiu2024richdreamer}, a sub-set of Objaverse \cite{deitke2023objaverse}.
Following the selection principle of \cite{li2024controllable}, we utilize the test samples collected by \cite{xu2024controlrm} gathered from 2 different datasets for zero-shot evaluation: 80 samples from Google Scanned Objects dataset (\textbf{GSO}) \cite{downs2022google}, and 80 samples from Amazon Berkeley Objects dataset (\textbf{ABO}) \cite{collins2022abo}.

\begin{table}[t]
	\centering
	\resizebox{1\linewidth}{!}{
		\begin{tabular}{l|ccc|ccc}
			\toprule
			\multirow{2}{*}{\textbf{Methods}}& \multicolumn{3}{c|}{\textbf{GOBJ$\rightarrow$GSO}}   & \multicolumn{3}{c}{\textbf{GOBJ$\rightarrow$ABO}}   \\
			 & \textbf{FID$^\downarrow$}    & \textbf{CLIP-I$^\uparrow$} & \textbf{CLIP-T$^\uparrow$} & \textbf{FID$^\downarrow$}    & \textbf{CLIP-I$^\uparrow$} & \textbf{CLIP-T$^\uparrow$} \\ \midrule \midrule
			\textbf{GSGEN \cite{chen2024text}}          & 344.61          & 0.740           & 0.289           & 366.47          & 0.669           & 0.259           \\
			\textbf{GaussianDreamer \cite{yi2024gaussiandreamer}}  & 278.70          & 0.810           & 0.300           & 225.38          & 0.787           & 0.277           \\
			\textbf{DreamGaussian \cite{tang2023dreamgaussian}} & 359.65          & 0.760           & 0.279           & 392.95          & 0.723           & 0.247           \\
			\textbf{VolumeDiffusion \cite{tang2023volumediffusion}}  & 299.61          & 0.715           & 0.259           & 350.46          & 0.679           & 0.242           \\
			\textbf{3DTopia \cite{hong20243dtopia}}    & 331.39          & 0.727           & 0.283           & 231.55          & 0.751           & 0.272           \\
			\textbf{MVControl \cite{li2024controllable}}  & 278.08          & 0.816           & 0.288           & 217.97          & 0.802           & 0.291           \\
			\textbf{ControLRM-T \cite{xu2024controlrm}} & 260.75          & \textbf{0.846}  & 0.289           & 202.14          & 0.827           & 0.282           \\
			\textbf{ControLRM-D \cite{xu2024controlrm}}& 171.13          & 0.838           & 0.302           & 181.84 & 0.836           & 0.292           \\ \midrule
			\textbf{CyC3D-T (Ours)}  & 255.30          & 0.840           & 0.295           & 196.55          & 0.836           & 0.291           \\
			\textbf{CyC3D-D (Ours)}     & \textbf{168.71} & \textbf{0.846}  & \textbf{0.303}  & \textbf{180.34}          & \textbf{0.843}  & \textbf{0.294}   \\ \bottomrule
		\end{tabular}
	}
	\vspace{-0.3cm}
	\caption{Generation quality (\textbf{FID}) and CLIP score (\textbf{CLIP-I}, \textbf{CLIP-T}) comparison with state-of-the-art controllable 3D generation methods \textbf{GSO}, and \textbf{ABO} benchmarks. $\uparrow$ means higher result is better, while $\downarrow$ means lower is better. The best results are emphasized in \textbf{bold}. We provide the average results of 4 different conditions (edge, sketch, depth, and normal) in the table.}
	\vspace{-0.6cm}
	\label{tab:quality}
\end{table}




\noindent\textbf{Training Details.}
In analogy with \cite{xu2024controlrm}, we initialize our network with the weights from pretrained OpenLRM-base \cite{hong2023lrm}.
For CyC3D-T, the conditional backbone is a transformer-based network; For CyC3D-D, the conditional backbone is a diffusion-based network.
The training might cost 2 days for CyC3D-T and 3 days for CyC3D-D on 32 Nvidia V100-32G GPUs.
More details are provided in the appendix.

\noindent\textbf{Evaluation and Metrics.}
For the task of controllable 3D generation task, we consider 4 different conditions: edge, sketch, depth, and normal.
For each condition, we evaluate the controllability by measuring the similarity between the input conditions and the extracted conditions from generated 3D contents following \cite{xu2024controlrm}.
For edge/sketch condition, we utilize \textbf{PSNR}, \textbf{SSIM}, and \textbf{MSE} as evaluation metrics;
For depth/normal condition, we utilize \textbf{MSE} to measure the similarity.
In analogy with \cite{li2025controlnet}, the large foundation models in monocular depth estimation are used to capture the depth priors for model-based evaluation.
If Midas \cite{ranftl2020towards} is used, the metric is noted as \textbf{M-MSE}.
If ZoeDepth \cite{bhat2023zoedepth} is used, the metric is noted as \textbf{Z-MSE}.
Following \cite{xu2024controlrm}, the depth consistency in 3D space compares the difference between the rendered depth map and the input condition depth map, noted as \textbf{R-MSE}.
In analogy with \cite{li2025controlnet}, the metric of \textbf{NB-MSE} uses the pre-trained model (Normal-BAE \cite{bae2021estimating}) in surface normal estimation to assess the normal consistency.
Furthermore, we also evaluate the depth-normal consistency between the rendered depth map and the input normal maps following \cite{xu2024controlrm}, noted as \textbf{DN-CON}.
Please refer to more details in the appendix.

\subsection{Experimental Results}




\begin{table}[!t]
\renewcommand\arraystretch{1.2}
\tabcolsep=3pt
\resizebox{1\linewidth}{!}{
\begin{tabular}{l|cccc|ccc}
\toprule
\multirow{2}{*}{\textbf{Methods}} & \multicolumn{4}{c|}{\textbf{Control}}                              & \multicolumn{3}{c}{\textbf{Quality}}             \\
                                  & \textbf{Edge$^\downarrow$} & \textbf{Sketch$^\downarrow$} & \textbf{Depth$^\downarrow$} & \textbf{Normal$^\downarrow$} & \textbf{FID$^\downarrow$} & \textbf{CLIP-I$^\uparrow$} & \textbf{CLIP-T$^\uparrow$} \\ \midrule \midrule
\textbf{ControlNet+OpenLRM-S \cite{hong2023lrm}}     & 0.1069        & 0.0459          & 0.1333         & 0.0255          & 268.7        & 0.807           & 0.292           \\
\textbf{ControlNet+OpenLRM-B \cite{hong2023lrm}}     & 0.0897        & 0.0452          & 0.1312         & 0.0206          & 258.04       & 0.822           & 0.293           \\
\textbf{ControlNet+OpenLRM-L \cite{hong2023lrm}}     & 0.0792        & 0.0447          & 0.1383         & 0.0226          & 256.43       & 0.824           & 0.294           \\
\textbf{ControlNet+TGS \cite{zou2024triplane}}           & 0.0523        & 0.0417          & 0.1403         & 0.0208          & 258.86       & 0.841           & 0.295           \\
\textbf{ControlNet+TripoSR \cite{tochilkin2024triposr}}       & 0.0978        & 0.0421          & 0.1188         & 0.0241          & 261.76       & 0.835           & 0.289           \\
\textbf{MVControlNet+LGM \cite{li2024controllable}}         & 0.0952        & 0.0370          & 0.0160         & 0.0315          & 278.08       & 0.816           & 0.288           \\ \hline
\textbf{CyC3D-T (Ours)}         & \textbf{0.0342}        & 0.0159          & 0.0065         & 0.0062          & 255.30       & 0.840           & 0.295           \\
\textbf{CyC3D-D (Ours)}         & 0.0347        & \textbf{0.0156}          & \textbf{0.0060}         & \textbf{0.0060}          & \textbf{168.71}       & \textbf{0.846}           & \textbf{0.303}           \\ \bottomrule
\end{tabular}
}
\vspace{-0.3cm}
\caption{Comparison with ControlNet-based baselines (ControlNet + Image-to-3D methods) on \textbf{GSO} benchmark.}
\vspace{-0.3cm}
\label{tab:controlnet}
\end{table}

\begin{table}[t]
\centering
\resizebox{0.8\linewidth}{!}{
\begin{tabular}{c|ccc|ccc}
\toprule
\multirow{2}{*}{\textbf{Loss}} & \multicolumn{3}{c|}{\textbf{Edge}}  & \multicolumn{3}{c}{\textbf{Sketch}} \\ & \textbf{PSNR} & \textbf{SSIM} & \textbf{MSE} & \textbf{PSNR} & \textbf{SSIM} & \textbf{MSE} \\ \midrule \midrule
\textbf{w/o $L_{\text{cond}}$}             & 12.87         & 0.841         & 0.0570       & 17.52         & 0.888         & 0.0187       \\
\textbf{w $L_{\text{cond}}$}             & 15.09         & 0.885         & 0.0347       & 18.38         & 0.890         & 0.0156       \\ \bottomrule
\end{tabular}
}
\vspace{-0.2cm}
\caption{Ablation experiments of 2D condition cycle consistency.}
\vspace{-0.6cm}
\label{tab:ablation:edge_sketch}
\end{table}

\begin{table}[t]
\centering
\renewcommand\arraystretch{1.2}
\resizebox{\linewidth}{!}{
\begin{tabular}{c|ccc|c|cc}
\toprule
\multirow{2}{*}{\textbf{Loss}} & \multicolumn{3}{c|}{\textbf{Depth}}               & \multirow{2}{*}{\textbf{Loss}} & \multicolumn{2}{c}{\textbf{Normal}} \\
                               & \textbf{M-MSE} & \textbf{Z-MSE} & \textbf{R-MSE} & & \textbf{NB-MSE}  & \textbf{DN-CON}  \\ \midrule \midrule
\textbf{w/o $L_{\text{cond}}$}             & 0.0383         & 0.0546         & 0.0650         & \textbf{w/o $L_{\text{cond}}$} & 0.0034           & 0.0205           \\
\textbf{w $L_{\text{cond}}$}             & 0.0192         & 0.0241         & 0.0321         & \textbf{w $L_{\text{cond}}$} & 0.0023           & 0.0187           \\
\textbf{w $L_{\text{cond}}+L_{\text{cond-d}}$}      & 0.0051         & 0.0041         & 0.0060         & \textbf{w $L_{\text{cond}}+L_{\text{cond-n}}$} & 0.0011           & 0.0056           \\ \bottomrule
\end{tabular}
}
\vspace{-0.2cm}
\caption{Ablation experiments of 3D condition cycle consistency.}
\vspace{-0.3cm}
\label{tab:ablation:depth_normal}
\end{table}

\noindent\textbf{Qualitative Results.}
We provide qualitative results of our CyC3D in Fig. \ref{fig:visualization}.
The backbone is inherited from ControLRM \cite{xu2024controlrm}, and the diffusion model is replaced with a pre-trained ControlNet. As the figure shows, our method not only performs well in generating high-quality 3D content but also in preserving details in conditional controls.
Furthermore, we also compare the qualitative results of our \name{} with other state-of-the-art methods, including ControLRM \cite{xu2024controlrm} and MVControl \cite{li2024controllable} in Fig. \ref{fig:qualitative}.

\noindent\textbf{Controllability Evaluation.}
The comparisons of controllability with state-of-the-art controllable 3D generation methods on \textbf{GSO}/\textbf{ABO} benchmarks are presented in Tab.  \ref{tab:controllability:gso}/\ref{tab:controllability:abo}.
Following \cite{xu2024controlrm}, we report the controllability metrics discussed in Sec.~\ref{sec:experiment:implementation} on all conditional controls.
From the tables, we can find that our \name{} can achieve state-of-the-art performance in the quantitative comparison with other baselines.
For example, the R-MSE of our CyC3D-D is 0.0060, which is much lower than the baselines of ControLRM-D (0.0650 R-MSE).
Furthermore, we also provide a qualitative comparison of controllability in Fig. \ref{fig:qualitative_controllability}.

\noindent\textbf{Generation Evaluation.}
To investigate whether improving controllability compromises generation quality, we provide the quantitative results of Fréchet Inception Distance (FID) \cite{heusel2017gans} and CLIP-score \cite{mohammad2022clip} following \cite{li2024controllable} in Tab. \ref{tab:quality}.
Furthermore, the qualitative results are provided in \fig \ref{fig:qualitative}.
We report the CLIP-I metric to quantify image similarity between reference and rendered images, and the CLIP-T metric to evaluate text-image alignment using textual annotations.
In rigorous zero-shot generalization experiments of cross-domain transitions towards GSO and ABO, our proposed CyC3D demonstrates statistically significant improvements over existing methods.
\noindent\textbf{Comparison with ControlNet-based Baselines.}
For a fair comparison, we also provide experimental results compared with ControlNet combined with image-to-3D reconstruction models.
The images produced by ControlNet are processed through image-to-3D reconstruction models to synthesize 3D assets, with foreground segmentation implemented using Rembg for background removal. The quantitative evaluation results on the GSO benchmark are presented in Tab.~\ref{tab:controlnet}, incorporating state-of-the-art image-to-3D systems including OpenLRM \cite{hong2023lrm}, TGS \cite{zou2024triplane}, and TripoSR \cite{tochilkin2024triposr} as comparative baselines.


\begin{figure}[t]
\centering
\includegraphics[width=\linewidth]{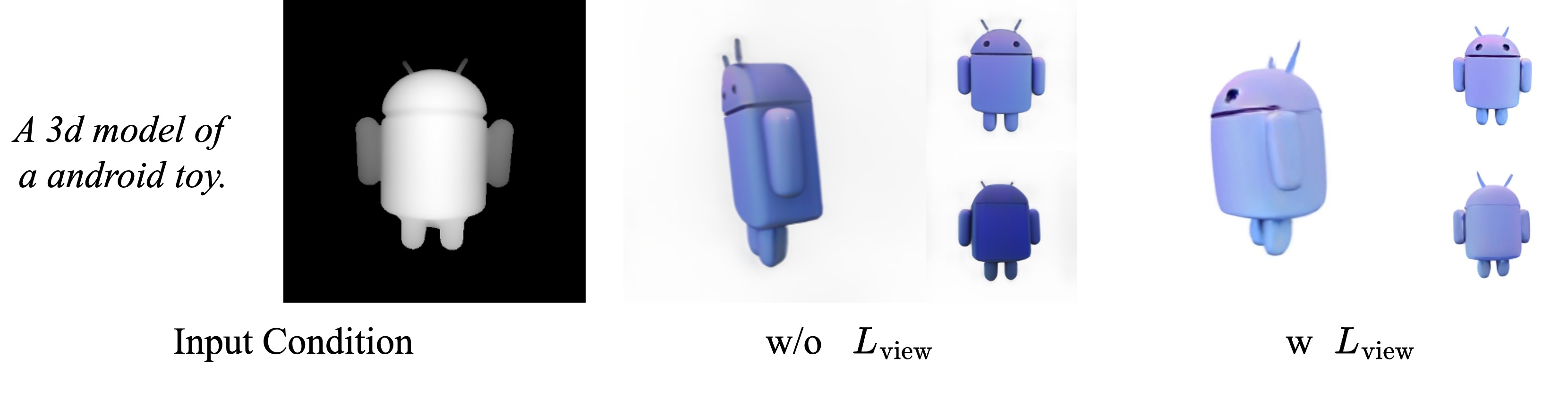}
\vspace{-0.8cm}
\caption{Ablation experiments of view cycle consistency.}
\vspace{-0.6cm}
\label{fig:ablation_view}
\end{figure}

\subsection{Ablation Study}

\noindent\textbf{Effect of Condition Cycle Consistency.}
To evaluate the effectiveness of 2D conditional control feedback (e.g., Eq. \ref{eq10} for edge/sketch), we provide ablation experiments in Tab. \ref{tab:ablation:edge_sketch}.
Furthermore, we also provide ablation results of 3D conditional control feedback (Eq. \ref{eq12} for depth, Eq. \ref{eq14} for normal) in Tab. \ref{tab:ablation:depth_normal}.
As the tables show, each term of condition cycle consistency can effectively improve the controllability upon different conditional controls.

\noindent\textbf{Effect of View Cycle Consistency.}
To evaluate the effectiveness of view cycle consistency feedback (e.g., Eq. \ref{eq17}), we provide qualitative ablation results in Fig. \ref{fig:ablation_view}.
As the figure shows, without view cycle consistency regularization, the generated 3D contents might be overfitting to the input viewpoint.
Though the rendered results in the input view are reasonable, the lack of regularization on the remaining views might lead to a low-quality 3D shape.

\section{Conclusion}

In this paper, we propose CyC3D, a novel framework for improving the controllability during 3D generation guided by different conditional controls (edge/sketch/depth/normal).
The core idea is the cycle-view consistency, which ensures accurate alignment between input conditions and generated models by rendering the generated 3D content from various viewpoints and re-extracting the conditions from these rendered views to feed back into the generation process. The quantitative and qualitative evaluation results on GSO and ABO benchmarks show the state-of-the-art controllability of CyC3D.

\bibliography{aaai2026}


\section*{Appendix}

Due to the page limit, we present additional details here as a comprehensive extension to the manuscript, including: 
\begin{enumerate}
    \item Detailed introduction of utilized training dataset, and the benchmark for evaluation. (Sec. \ref{suppl:dataset})
    \item Detailed explanation of the evaluation metric on controllability. (Sec. \ref{suppl:metric})
    \item Detailed explanation of the evaluation metric on generation quality. (Sec. \ref{suppl:metric_qual})
    \item Details for reproducing the baseline results used in the paper. (Sec. \ref{suppl:baseline})
    \item Implementation details of our proposed CyC3D. (Sec. \ref{suppl:implementation})
    \item Additional experimental results including quantitative and qualitative results. (Sec. \ref{suppl:experiments})
    \item Discussion about failure cases and limitations of the proposed method. (Sec. \ref{suppl:failure})
\end{enumerate}

\section*{Utilized Datasets}
\label{suppl:dataset}

\noindent\underline{\textbf{Training Dataset}}
Instead of using the full version of Objaverse \cite{deitke2023objaverse} as the training dataset, we utilize G-buffer Objaverse (G-Objaverse) \cite{qiu2024richdreamer} as the training dataset.
The original Objaverse contains about 800K 3D objects, while the G-Objaverse contains about 280K 3D objects.
Same as controllable 2D generation (i.e. ControlNet \cite{zhang2023adding}), the training does not have to be the full Objaverse dataset, our method can work on a smaller training dataset (G-Objaverse).
In G-Objaverse, each 3D object is rendered to RGB, Depth, and Normal maps on 40 different viewpoints, including 24 views at elevation range from $5^{\circ}$ to $30^{\circ}$, rotation $=\{ r \times 15^{\circ} | r \in [0, 23] \}$, and 12 views at elevation from $-5^{\circ}$ to $5^{\circ}$, rotation $= \{ r \ times 30^{\circ} | r \in [0, 11] \}$, and 2 views for top and bottom respectively.
The rendered RGB images on 40 different views are used in the training dataset.
The corresponding text caption of each 3D object is also provided by Cap3D \cite{luo2024scalable}.
After filtering the unavailable samples, we can obtain a training set with 260K samples and a test set with 2800 samples.

\noindent\underline{\textbf{Evaluation Dataset of GOBJ}}
For evaluation, 118 samples in the separated test set of G-Objaverse are selected following \cite{xu2024controlrm}, which are absent from the training data.
The samples are collected following selection principles noted in MVControl \cite{li2024controllable} and TripoSR \cite{tochilkin2024triposr}.
The text prompts are from Cap3D \cite{luo2024scalable}.
One reference view is manually selected from all 40 available views and the condition map of edge/sketch/depth/normal is extracted from the reference image.
The reference condition map is treated as input, and the remaining images are treated as ground truth images for evaluation.
This evaluation set is noted as \textbf{GOBJ}.

\noindent\underline{\textbf{Evaluation Dataset of GSO}}
Furthermore, 80 samples of the Google Scanned Objects dataset \cite{downs2022google} are selected following \cite{xu2024controlrm} for zero-shot evaluation.
This dataset features more than one thousand 3D-scanned household items serving as a valuable resource for assessing the zero-shot generalization abilities of the proposed method.
In analogy with \textbf{GOBJ}, we manually select the reference view from the overall 32 available views provided in the dataset.
The condition map of edge/sketch/depth/normal is extracted from the reference view image.
The text prompts are obtained via captioning the reference image with BLIP2 \cite{li2023blip}.
The condition map and the text prompt of each  sample are input to the model, and the remaining views are utilized as the ground truth images for evaluation.
This evaluation set is noted as \textbf{GSO}.

\noindent\underline{\textbf{Evaluation Dataset of ABO}}
For zero-shot evaluation, we also select 80 samples of Amazon Berkeley Objects dataset \cite{collins2022abo} following \cite{xu2024controlrm}.
The Amazon Berkeley Objects dataset is a comprehensive 3D dataset comprising product catalog images, metadata, and artist-designed 3D models featuring intricate geometries and materials based on real household objects.
In analogy with \textbf{GSO}, we manually select the reference view from the overall 72 views provided in the dataset.
The condition map of edge/sketch/depth/normal is extracted from the selected reference view image.
The text prompts are generated via captioning the reference image with BLIP2 \cite{li2023blip}.
The condition map and the text prompt of each sample are input to the model, and the remaining views are utilized as the ground truth images for evaluation.
This evaluation set is noted as \textbf{ABO}.

\section*{Evaluation Metric on Controllability}
\label{suppl:metric}

In this section, we introduce the details of the evaluation metrics of controllability on different conditional controls.

Assume that the generated 3D content is $P$, and the camera transformation matrix is $T_r$, the rendered image and depth via volume rendering \cite{mildenhall2021nerf} can be written as follows:
\begin{equation}
\begin{aligned}
    I_{r} &= f_{\text{render}}^I (P, T_r) \\
    D_{r} &= f_{\text{render}}^D (P, T_r)
\end{aligned}
\end{equation}
where $I_{r}$ is the rendered image at viewpoint $r$, and $D_{r}$ is the rendered depth at viewpoint $r$.
$f_{\text{render}}^I$ is the volume rendering function of the RGB image based on color and density.
$f_{\text{render}}^D$ is the volume rendering function of the depth map based on density only.

To evaluate the controllability between rendered results and input condition maps, we need a corresponding condition extractor $f_{\text{cond}}$ according to each of the conditional controls:
\begin{equation}
    f_{\text{cond}} = 
    \left\{
    \begin{aligned}
        & f_{\text{canny}}, \;\; \text{if} \;\;  C_v = C_e  \; (\text{edge condition}) \\
        & f_{\text{sketch}}, \;\; \text{if} \;\;  C_v = C_s  \; (\text{sketch condition}) \\
        & f_{\text{depth}}, \;\; \text{if} \;\;  C_v = C_d  \; (\text{depth condition}) \\
        & f_{\text{normal}}, \;\; \text{if} \;\;  C_v = C_n  \; (\text{normal condition}) 
    \end{aligned}
    \right.
\end{equation}
where $f_{\text{canny}}$, $f_{\text{sketch}}$, $f_{\text{depth}}$, $f_{\text{normal}}$ are respectively the condition extractor of under edge/sketch/depth/normal controls.
$C_v$ is the input condition.
$C_e$/$C_s$/$C_d$/$C_n$ respectively represents the edge/sketch/depth/normal condition map.

A brief summarization of controllability metrics on each condition is shown as follows:
\begin{itemize}
    \item Edge condition: \textbf{PSNR}, \textbf{SSIM}, \textbf{MSE}. (Sec. \ref{suppl:metric:edge_sketch})
    \item Sketch condition: \textbf{PSNR}, \textbf{SSIM}, \textbf{MSE}. (Sec. \ref{suppl:metric:edge_sketch})
    \item Depth condition: \textbf{M-MSE}, \textbf{Z-MSE}, \textbf{R-MSE}. (Sec. \ref{suppl:metric:depth})
    \item Normal condition: \textbf{NB-MSE}, \textbf{DN-CON}. (Sec. \ref{suppl:metric:normal})
\end{itemize}

\subsection*{Controllability on Edge/Sketch}
\label{suppl:metric:edge_sketch}

To evaluate the controllability of visual conditional controls like edge or sketch, we modify the widely used metric of \textbf{PSNR}, \textbf{SSIM}, \textbf{MSE} to measure the difference between the condition maps of the input sample and generated sample.
Note that $f_{\text{cond}} = f_{\text{canny}}$ if the conditional control is edge, and $f_{\text{cond}} = f_{\text{sketch}}$ if the conditional control is sketch.
$f_{\text{canny}}$ is the Canny descriptor \cite{canny1986computational} which extracts edge maps from the input images.
We share the same hyper-parameter setting of $f_{\text{canny}}$ as ControlNet \cite{zhang2023adding}.
$f_{\text{sketch}}$ is a pre-trained sketch extraction network provided by ControlNet \cite{zhang2023adding}.
$C_{\text{cond}}$ is the input conditional map at viewpoint $r$.
$C_{\text{cond}} = C_e$ if the conditional control is edge, and $C_{\text{cond}} = C_s$ if the conditional control is sketch.

In the following, we provide the definition of the quantitative metrics for evaluating the controllability:

\noindent\underline{\textbf{PSNR}}
Given the camera pose $T_r$ at viewpoint $r$, the Peak signal-to-noise ratio (PSNR) is calculated between the condition map of the rendered image in the generated 3D content, and the one of the ground truth image $I_r^{\text{gt}})$ at viewpoint $r$:
\begin{equation}
    \small
    \text{PSNR}_{\text{cond}} (P, T_r, I_r^{\text{gt}}) = \text{PSNR} (f_\text{cond} (f_{\text{render}}^I (P, T_r)), C_\text{cond})
\end{equation}

\noindent\underline{\textbf{SSIM}}
Given the camera pose $T_r$ at viewpoint $r$, the Structure Similarity Index Measure (SSIM) is calculated between the condition map of the rendered image in the generated 3D content, and the one of the ground truth image $I_r^{\text{gt}})$ at viewpoint $r$:
\begin{equation}
    \small
    \text{SSIM}_{\text{cond}} (P, T_r, I_r^{\text{gt}}) = \text{SSIM} (f_\text{cond} (f_{\text{render}}^I (P, T_r)), C_\text{cond})
\end{equation}

\noindent\underline{\textbf{MSE}}
Given the camera pose $T_r$ at viewpoint $r$, the Mean Squared Error (MSE) difference is calculated between the condition map of the rendered image in the generated 3D content, and the one of the ground truth image $I_r^{\text{gt}})$ at viewpoint $r$:
\begin{equation}
    \small
    \text{MSE}_{\text{cond}} (P, T_r, I_r^{\text{gt}}) = \text{MSE} (f_\text{cond} (f_{\text{render}}^I (P, T_r)), C_\text{cond})
\end{equation}

\subsection*{Controllability on Depth}
\label{suppl:metric:depth}

To evaluate the controllability of depth conditional control, we build the metrics in two ways:

(1) \textbf{Model-based metric.} The consistency of depth maps estimated by large pre-trained models such as Midas \cite{ranftl2020towards} and Zoe-Depth \cite{bhat2023zoedepth} can reflect the depth priors of the images.
Following ControlNet++ \cite{li2025controlnet}, we build the \textbf{M-MSE} and \textbf{Z-MSE} metrics to evaluate the depth consistency between the input condition and the generated one.
If Midas model \cite{ranftl2020towards} is used, $f_{\text{cond}} = f_{\text{midas}}$.
If Zoe-Depth \cite{bhat2023zoedepth} is used, $f_{\text{cond}} = f_{\text{zoe-depth}}$.
The input conditional map $C_{\text{cond}}$ at viewpoint $r$ is $C_d$.
The corresponding ground truth image is noted as $I_r^{\text{gt}}$.

\noindent\underline{\textbf{M-MSE}}
This metric firstly extracts the relative depth maps from the images via pre-trained Midas model $f_\text{midas}$.
Then, the MSE difference between the estimated depth maps is calculated as follows:
\begin{equation}
    \small
    \text{M-MSE} (P, T_r, I_r^{\text{gt}}) = \text{MSE} (f_\text{midas} (f_{\text{render}}^I (P, T_r)), f_\text{midas}(I_r^{\text{gt}}))
\end{equation}

\noindent\underline{\textbf{Z-MSE}}
This metric firstly extracts the relative depth maps from the images via pre-trained Zoe-Depth model $f_{\text{zoe-depth}}$.
Then, the MSE difference between the estimated depth maps is computed as follows:
\begin{equation}
    \small
    \text{Z-MSE} (P, T_r, I_r^{\text{gt}}) = \text{MSE} (f_\text{zoe-depth} (f_{\text{render}}^I (P, T_r)), f_\text{zoe-depth}(I_r^{\text{gt}}))
\end{equation}

(2) \textbf{3D-based metric.}
Since the model-based metrics of M-MSE and Z-MSE only capture the depth consistencies from rendered 2D images, it is necessary to evaluate the consistency between the generated 3D representation and the 3D conditional control (depth).
In consequence, the 3D-based metrics of \textbf{R-MSE} measures the MSE difference between the rendered depth map of the generated 3D content $P$ at viewpoint $r$ and the input depth condition map $C_d$.

\noindent\underline{\textbf{R-MSE}}
This metric compares the difference between the rendered depth map and the input depth condition map as follows:
\begin{equation}
    \small
    \text{R-MSE} (P, T_r, C_d) = \text{MSE} (f_{\text{norm}} (f_{\text{render}}^D (P, T_r)), C_d)
\end{equation}
where $f_{\text{norm}}$ is the normalization function.
Since the rendered depth map and the input depth condition map have different scales, it is necessary to align the scale of these depth maps.
Assume that the input depth condition map is already normalized to a range of 0 to 1, we can also normalize the rendered depth map to the same range of 0 to 1 via $f_{\text{norm}}$.

\subsection*{Controllability on Normal}
\label{suppl:metric:normal}

To evaluate the controllability of normal conditional control, we build the metrics in two ways:

(1) \textbf{Model-based metric}.
The consistency of normal maps estimated by pre-trained models like Normal-BAE \cite{bae2021estimating} can reflect the normal priors of the images.
Following ControlNet++ \cite{li2025controlnet}, we build \textbf{NB-MSE} metric to evaluate the normal consistency between the input normal condition and the one extracted from the generated result.
The input normal condition is noted as $C_n$, and the ground truth image at viewpoint $r$ is noted as $I_r^{\text{gt}}$.
We utilize the pre-trained model of Normal-BAE provided by ControlNet \cite{zhang2023adding}, the condition extractor is: $f_{\text{cond}} = f_{\text{nb-mae}}$.

\noindent\underline{\textbf{NB-MSE}}
This metric captures the similarity of normal priors captured by pre-trained model of Normal-BAE.
\begin{equation}
    \small
    \text{NB-MSE} (P, T_r, I_r^{\text{gt}}) = \text{MSE} (f_\text{nb-mae} (f_{\text{render}}^I (P, T_r)), f_\text{nb-mae}(I_r^{\text{gt}}))
\end{equation}

(2) \textbf{3D-based metric}.
Besides the model-based metric, we further evaluate the depth-normal consistency between the directly rendered depth map and the input normal condition.
It can capture the consistency in 3D space directly.
Specifically, we utilize the metric of \textbf{DN-CON} to measure the depth-normal consistency based on the orthogonality between normal and local surface tangent \cite{yang2018unsupervised}.

\noindent\underline{\textbf{DN-CON}}
The rendered depth is converted to a normalized normal maps via a differentiable module \cite{huang2021m3vsnet}, and be compared with the input normal condition as follows:
\begin{equation}
    \small
    \text{DN-CON} (P, T_r, C_n) = \text{MSE} (f_{\text{d2n}} (f_{\text{render}}^D (P, T_r)), C_n)
\end{equation}




\section*{Evaluation Metric on Generation Quality}
\label{suppl:metric_qual}

In the main manuscript, we also provide the quantitative comparison with other methods.
Here we introduce the details of related metrics:

\noindent\underline{\textbf{FID}}
In analogy with LATTE3D \cite{xie2024latte3d}, we compute the Fréchet Inception Distance (FID) \cite{heusel2017gans} between the rendered images of the generated 3D object and the collected ground truth multi-view images in the evaluation benchmarks.
This metric can reflect how well the generation results algin with the ground truth multi-view images in visual quality.

\noindent\underline{\textbf{CLIP-I}}
Following MVControl \cite{li2024controllable}, we also compute the CLIP score of image features extracted from the renderings of the generated 3D content and the collected ground truth images, noted as CLIP-I.
This metric aims to measure the similarity between the rendering results of generated 3D objects and the ground truth images.

\noindent\underline{\textbf{CLIP-T}}
Following MVControl \cite{li2024controllable}, we also measure the CLIP score between image features extracted from the renderings and the text features extracted from the given text prompt.
This metric aims to reveal the similarity between the generated 3D contents and the given text descriptions.

\noindent\underline{\textbf{View Settings}}
The view setting of evaluation protocol in MVControl \cite{li2024controllable} only compute the metrics between the generated multi-view images and the ground turth image on the reference view.
However, merely evaluating the performance on singe one view is not comprehensive for evaluating the 3D generated contents.
In consequence, we utilize the following settings to build more comprehensive metrics:
(1) \textbf{All Views}: All views are taken into account when calculating the FID, CLIP-I, and CLIP-T;
(2) \textbf{Front-K Views}: Given the provision of only one reference view, the views on the back side may lack crucial cues for precise prediction, potentially leading to unreliable results in multi-view scenarios. 
Therefore, incorporating an additional evaluation of the views in front of the reference view is necessary. 
Consequently, we select the K=4 views closest to the given reference view for further metric computation.

\section*{Baselines for Comparisons}
\label{suppl:baseline}

For comparison, we select a series of state-of-the-art methods as baselines in the controllable 3D generation task, including:
\begin{enumerate}
    \item \textbf{Optimization-based methods:} GSGEN \cite{chen2024text}, GaussianDreamer \cite{yi2024gaussiandreamer}, and DreamGaussian \cite{tang2023dreamgaussian}. These optimization-based methods iteratively distill the prior of large pre-trained generative diffusion models to generate the corresponding 3D contents.
    \item \textbf{Feed-forward-based methods:} VolumeDiffusion \cite{tang2023volumediffusion}, 3DTopia \cite{hong20243dtopia}. These models directly build a diffusion model on 3D representation like Triplane NeRF \cite{mildenhall2021nerf}, and generate 3D contents via the guidance of the input condition on the diffusion model.
    \item \textbf{Controllable 3D generation methods:} MVControl \cite{li2024controllable}, and ControLRM \cite{xu2024controlrm}. These models are state-of-the-art models built for controllable 3D generation.
    They accept conditional maps and text prompts as input, and generate corresponding 3D contents via the pre-trained model.
\end{enumerate}

\noindent
Below we provide the details needed to reproduce the baseline methods shown in the paper.

\noindent\underline{\textbf{GSGEN}}
We use the \footnote{https://github.com/gsgen3d/gsgen}{official code} provided by GSGEN \cite{chen2024text}.
The text prompts in the evaluation benchmarks are used as the input to generate the 3D contents.
The default hyper-parameters of GSGEN are used.

\noindent\underline{\textbf{GaussianDreamer}}
We use the \footnote{https://github.com/hustvl/GaussianDreamer}{official code}  provided by the GaussianDreamer \cite{yi2024gaussiandreamer}.
The text prompts in the evaluation benchmarks are set as input to generate the 3D assets.
The default hyper-parameters in their official implementation are used.

\noindent\underline{\textbf{DreamGaussians}}
We use the \footnote{https://github.com/dreamgaussian/dreamgaussian}{official code} of DreamGaussian \cite{tang2023dreamgaussian}.
The text prompts in the evaluation benchmarks are used as the input for generating the 3D contents.
The default hyper-parameters of their official implementation are directly used for experiments.

\noindent\underline{\textbf{VolumeDiffusion}}
We use the \footnote{https://github.com/PeterZs/VolumeDiffusion}{official code} of VolumeDiffusion \cite{tang2023volumediffusion}.
The text prompts of the evaluation datasets are used as input.
We use the \footnote{https://github.com/PeterZs/VolumeDiffusion?tab=readme-ov-file\#inference}{pre-trained model} of VolumeDiffusion to generate the 3D contents.

\begin{figure*}[t]
    \centering
    \includegraphics[width=\linewidth]{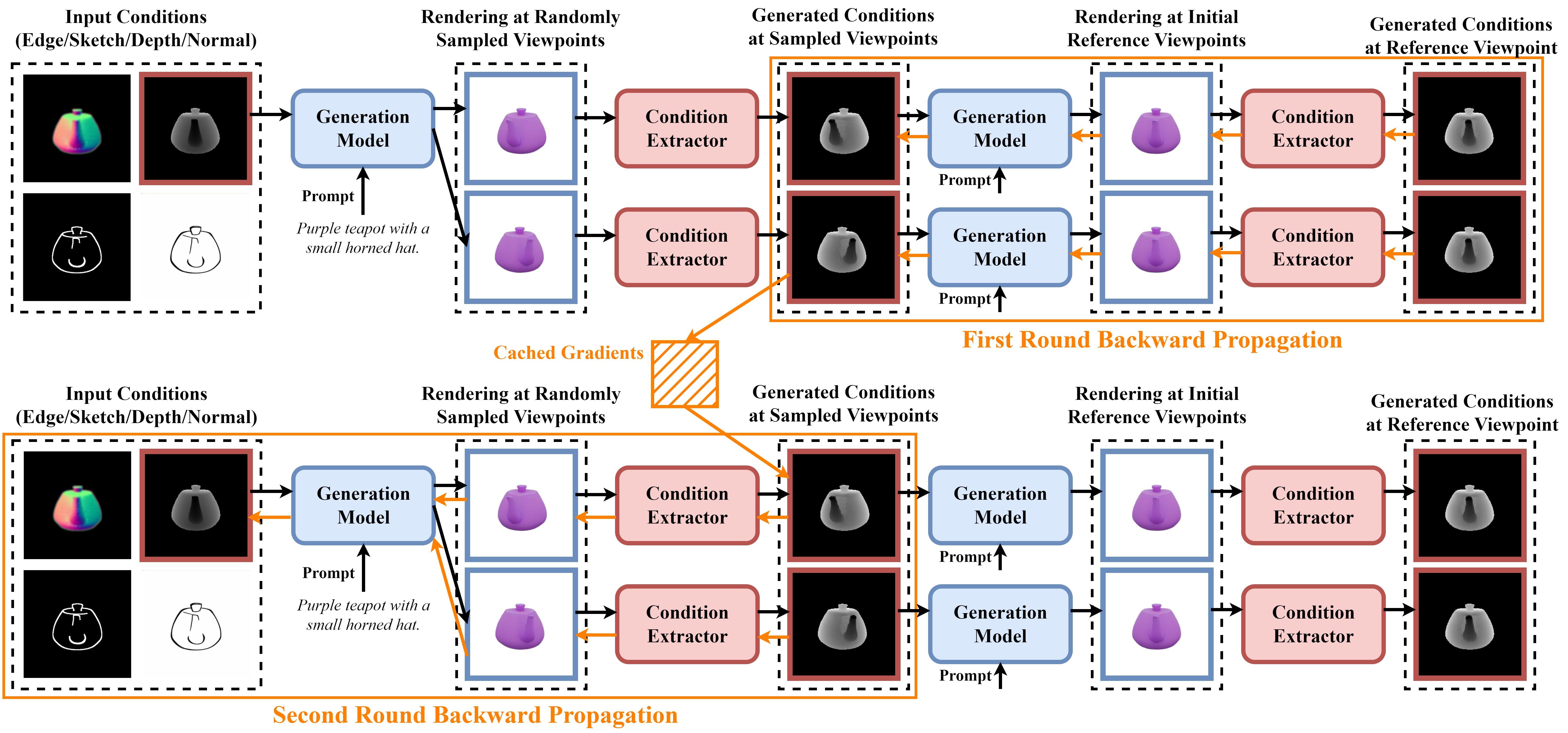}
    \vspace{-0.6cm}
    \caption{Illustration of Pseudo Cycle Back-Propagation for implementing the cycle consistency with less GPU memory costs.}
    \label{fig:pseudo_cycle_back_propagation}
\end{figure*}

\begin{table*}[t]
\centering
\resizebox{\linewidth}{!}{
\begin{tabular}{l|ccc|ccc|ccc}
\toprule
\multirow{2}{*}{\textbf{Methods}} & \multicolumn{3}{c|}{\textbf{GOBJ$\rightarrow$GOBJ}} & \multicolumn{3}{c|}{\textbf{GOBJ-$\rightarrow$GSO}} & \multicolumn{3}{c}{\textbf{GOBJ$\rightarrow$ABO}} \\
                                  & \textbf{FID$^\downarrow$}     & \textbf{CLIP-I$^\uparrow$} & \textbf{CLIP-T$^\uparrow$} & \textbf{FID$^\downarrow$}    & \textbf{CLIP-I$^\uparrow$} & \textbf{CLIP-T$^\uparrow$} & \textbf{FID$^\downarrow$}    & \textbf{CLIP-I$^\uparrow$} & \textbf{CLIP-T$^\uparrow$} \\ \midrule
\textbf{GSGEN \cite{chen2024text}}                    & 340.39           & 0.762           & 0.298           & 344.61          & 0.740           & 0.289           & 366.47          & 0.669           & 0.259           \\
\textbf{GaussianDreamer \cite{yi2024gaussiandreamer}}          & 268.95           & 0.803           & \textbf{0.309}  & 278.7           & 0.810           & 0.300           & 225.38          & 0.787           & 0.277           \\
\textbf{DreamGaussian \cite{tang2023dreamgaussian}}           & 351.87           & 0.761           & 0.279           & 359.65          & 0.760           & 0.279           & 392.95          & 0.723           & 0.247           \\
\textbf{VolumeDiffusion \cite{tang2023volumediffusion}}          & 327.76           & 0.725           & 0.241           & 299.61          & 0.715           & 0.259           & 350.46          & 0.679           & 0.242           \\
\textbf{3DTopia \cite{hong20243dtopia}}                  & 289.02           & 0.749           & 0.280           & 331.39          & 0.727           & 0.283           & 231.55          & 0.751           & 0.272           \\
\textbf{MVControl \cite{li2024controllable}}                & 251.71           & 0.811           & 0.291           & 278.08          & 0.816           & 0.288           & 217.97          & 0.802           & 0.291           \\
\textbf{ControLRM-T \cite{xu2024controlrm}}              & 166.03           & 0.879           & 0.292           & 260.75          & \textbf{0.846}  & 0.289           & 202.14          & 0.827           & 0.282           \\
\textbf{ControLRM-D \cite{xu2024controlrm}}              & \textbf{163.25}  & \textbf{0.887}  & 0.300           & 171.13          & 0.838           & 0.302           & 181.84 & 0.836           & 0.292           \\ \midrule
\textbf{CyC3D-T (Ours)}              & 170.84           & 0.880           & 0.292           & 255.30          & 0.840           & 0.295           & 196.55          & 0.836           & 0.291           \\
\textbf{CyC3D-D (Ours)}              & 171.21           & 0.881           & 0.297           & \textbf{168.71} & \textbf{0.846}  & \textbf{0.303}  & \textbf{180.34}          & \textbf{0.843}  & \textbf{0.294}  \\ \bottomrule
\end{tabular}
}
  
\caption{Generation quality (\textbf{FID}) and CLIP score (\textbf{CLIP-I}, \textbf{CLIP-T}) comparison under \textbf{All View} setting with other state-of-the-art controllable 3D generation method on \textbf{GOBJ}, \textbf{GSO}, and \textbf{ABO} benchmarks. $\uparrow$ means higher result is better, while $\downarrow$ means lower is better.}
\label{tab:generation_quality_all_view}
\end{table*}

\begin{table*}[t]
\centering
\resizebox{\linewidth}{!}{
\begin{tabular}{l|ccc|ccc|ccc}
\toprule
\multirow{2}{*}{\textbf{Methods}} & \multicolumn{3}{c|}{\textbf{GOBJ$\rightarrow$GOBJ}} & \multicolumn{3}{c|}{\textbf{GOBJ$\rightarrow$GSO}} & \multicolumn{3}{c}{\textbf{GOBJ$\rightarrow$ABO}} \\
                                  & \textbf{FID$^\downarrow$}     & \textbf{CLIP-I$^\uparrow$} & \textbf{CLIP-T$^\uparrow$} & \textbf{FID$^\downarrow$}    & \textbf{CLIP-I$^\uparrow$} & \textbf{CLIP-T$^\uparrow$} & \textbf{FID$^\downarrow$}    & \textbf{CLIP-I$^\uparrow$} & \textbf{CLIP-T$^\uparrow$} \\ \midrule
\textbf{GSGEN \cite{chen2024text}}                    & 357.13           & 0.781           & 0.310           & 360.57          & 0.759           & 0.300           & 376.56          & 0.691           & 0.272           \\
\textbf{GaussianDreamer \cite{yi2024gaussiandreamer}}          & 282.59           & 0.823           & 0.321           & 287.20          & 0.829           & 0.311           & 226.56          & 0.831           & 0.306           \\
\textbf{DreamGaussian \cite{tang2023dreamgaussian}}           & 368.76           & 0.783           & 0.293           & 373.53          & 0.784           & 0.290           & 406.35          & 0.750           & 0.290           \\
\textbf{VolumeDiffusion \cite{tang2023volumediffusion}}          & 348.39           & 0.752           & 0.257           & 316.35          & 0.742           & 0.273           & 372.49          & 0.715           & 0.262           \\
\textbf{3DTopia \cite{hong20243dtopia}}                  & 329.29           & 0.808           & 0.312           & 369.27          & 0.799           & 0.311           & 259.88          & 0.844           & 0.312           \\
\textbf{MVControl \cite{li2024controllable}}                & 280.40           & 0.856           & 0.318           & 301.31          & 0.870           & 0.312           & 236.81          & 0.868           & 0.316           \\
\textbf{ControLRM-T \cite{xu2024controlrm}}              & \textbf{144.02}  & 0.932           & 0.323           & 251.57          & 0.912           & 0.316           & 160.12          & 0.915           & 0.320           \\
\textbf{ControLRM-D \cite{xu2024controlrm}}              & 148.76           & \textbf{0.935}  & \textbf{0.330}  & 247.06          & 0.908           & 0.322           & \textbf{152.37} & \textbf{0.918}  & \textbf{0.324}  \\ \midrule
\textbf{CyC3D-T (Ours)}              & 155.08           & 0.930           & 0.322           & 247.03          & 0.913           & 0.320           & 159.03          & 0.904           & 0.322           \\
\textbf{CyC3D-D (Ours)}              & 155.82           & 0.931           & 0.327           & \textbf{215.98} & \textbf{0.915}  & \textbf{0.324}  & 154.85          & 0.908           & \textbf{0.324} \\ \bottomrule
\end{tabular}
}
\caption{Generation quality (\textbf{FID}) and CLIP score (\textbf{CLIP-I}, \textbf{CLIP-T}) comparison under \textbf{Front-K View} setting with other state-of-the-art controllable 3D generation method on \textbf{GOBJ}, \textbf{GSO}, and \textbf{ABO} benchmarks. $\uparrow$ means higher result is better, while $\downarrow$ means lower is better.}
\label{tab:generation_quality_front_k_view}
\end{table*}

\noindent\underline{\textbf{3DTopia}}
We use the \footnote{https://github.com/3DTopia/3DTopia}{official code} of 3DTopia \cite{hong20243dtopia}.
The text prompts in the evaluation datasets are used as input.
We use their open-sourced \footnote{https://huggingface.co/hongfz16/3DTopia}{pre-trained checkpoints} for experiments.
The default parameters in the official code are used.

\noindent\underline{\textbf{MVControl}}
We use the \footnote{https://github.com/WU-CVGL/MVControl}{official implementation} of the state-of-the-art controllable 3D generation model, MVControl \cite{li2024controllable}.
The open-source pre-trained checkpoints of different conditional controls are used for experiments, including: \footnote{https://huggingface.co/lzq49/mvcontrol-4v-canny}{Edge (Canny)}, \footnote{https://huggingface.co/lzq49/mvcontrol-4v-depth}{Depth}, \footnote{https://huggingface.co/lzq49/mvcontrol-4v-normal}{Normal}, \footnote{https://huggingface.co/lzq49/mvcontrol-4v-scribble}{Sketch}.
The condition map on the reference view and the corresponding text prompt are fed to the model as input.
The same parameters in the official code are used for a fair comparison.

\noindent\underline{\textbf{ControLRM}}
We use the \footnote{https://toughstonex.github.io/controlrm.github.io/}{official implementation} of ControLRM \cite{xu2024controlrm}.
the open-source pre-trained checkpoints of different conditional controls are used for experiments, including: \footnote{https://toughstonex.github.io/controlrm.github.io/}{Edge (Canny)}, \footnote{https://toughstonex.github.io/controlrm.github.io/}{Depth}, \footnote{https://toughstonex.github.io/controlrm.github.io/}{Normal}, \footnote{https://toughstonex.github.io/controlrm.github.io/}{Sketch}.
The condition map on the reference view and the corresponding text prompt are fed to the model as input.
The same parameters in the official code are used for a fair comparison.

\section*{Implementation Details}
\label{suppl:implementation}

\noindent\underline{\textbf{Backbone Network}}
In analogy with \cite{xu2024controlrm}, we initialize our network with the weights from pretrained OpenLRM-base \cite{hong2023lrm}.
The image-conditioned transformer is removed and replaced with our customized conditional backbone network taking visual conditions (edge/sketch/depth/normal) as input.
The output feature is fed to the activated cross-attention layers in the triplane transformer of OpenLRM, while the remaining parameters are kept frozen.
ControLRM \cite{xu2024controlrm} has 2 variations with different conditional backbones: ControLRM-T with transformer backbone, and ControLRM-D with diffusion backbone.
In analogy, we inherit the same conditional backbone of ControLRM to our CyC3D, and note them as CyC3D-T/-D rescpectively.

\noindent\underline{\textbf{Training Details}}
We utilize 32 Nvidia V100-32G GPUs for training CyC3D-T and CyC3D-D.
For CyC3D-T, the training might cost 2 days.
For CyC3D-D, the training might cost 3 days.
We utilize the AdamW optimizer with a learning rate of 4e-4.
The resolution of the condition map (edge/sketch/depth/normal) is set to $336 \times 336$.
To save GPU memories, the resolution of the rendered image is set to $256 \times 256$ in default.
The training code is heavily based on the open-source project \footnote{https://github.com/3DTopia/OpenLRM}{OpenLRM}, and will be release in the future.

\noindent\underline{\textbf{Pseudo Cycle Back-Propagation}}
In the implementation, a non-negligible issue is the huge memory cost during training.
The volume rendering has to be run twice in the cycle consistency, hence even the V100-32G GPU might suffer from the problem of GPU memory overflow.
To handle this problem, we propose \textbf{Pseudo Cycle Back-Propagation} to trade space with time, as shown in Fig. \ref{fig:pseudo_cycle_back_propagation}.
The orange arrows in the figure denotes the back-ward gradients in the computation graph.
Instead of calculating the whole computation graph in one round, we separate it into two rounds as the figure shows.
In the first round, we back-propagate the gradients to the intermediate output (\emph{Generated conditions at randomly sampled views}).
The gradients on the intermediate output are cached and saved as tensors.
In the second round, we back-propagate these cached gradients to the corresponding nodes in the computation graph, and finish the remaining process of back-propagation.
In summary, we replace the original computation pipeline (one forward-propagation, one backward-propagation) with our modified computation pipeline (one forward-propagation, two backward-propagation).
It can effectively reduce the GPU memory cost during the training phase.

\section*{Additional Experimental Results}
\label{suppl:experiments}

\subsection*{Time Efficiency}

\begin{figure}[t]
    \centering
    \includegraphics[width=\linewidth]{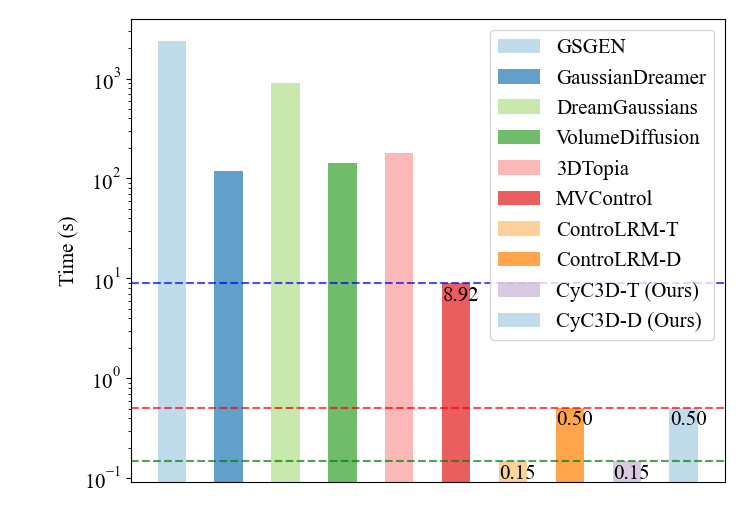}
    \caption{Time efficiency comparison with other methods in controllable 3D generation.}
    \label{fig:time_efficiency}
\end{figure}

In Fig. \ref{fig:time_efficiency}, we compare our CyC3D with other methods in time efficiency.
The inference time of the model on a single V100-32G GPU is reported in the figure.
Since our model utilizes the same backbone as ControLRM \cite{xu2024controlrm}, our proposed method will not add extra inference budget to the backbone network meantime improving the controllability quality.

\subsection*{Additional Quantitative Results on Generation Quality}

The controllable 3D generation model should not only improve the controllability, but also maintain a good generation quality.
Usually, there exists a trade-off between these two aspects depending on which property one prefers.
In consequence, we also provide the quantitative comparison of the generation quality here.
As discussed in Sec. \ref{suppl:metric_qual}, we compare different method on \textbf{FID}/\textbf{CLIP-I}/\textbf{CLIP-T} under two different view settings (\textbf{All View}/\textbf{Front-K View}).
The quantitative results under \textbf{All View} setting are presented in Tab. \ref{tab:generation_quality_all_view}.
The quantitative results under \textbf{Front-K Views} setting are shown in Tab. \ref{tab:generation_quality_front_k_view}.
Our model can maintain competitive and even better generation quality under zero-shot generalization evaluation, i.e. \textbf{GOBJ$\rightarrow$GSO}, \textbf{GOBJ$\rightarrow$ABO}.
But the generation quality on the same dataset might be slightly worse than the existing SOTA.
This implies that our model can generalize to unseen domains well by sacrificing the over-fitting effect to the training domain.

\subsection*{Comparison under Different Conditional Controls}

\begin{table}[t]
\centering
\resizebox{\linewidth}{!}{
\begin{tabular}{c|l|cccc|c}
\toprule
\textbf{Metrics}                  & \textbf{Methods}     & \textbf{Edge} & \textbf{Depth} & \textbf{Normal} & \textbf{Sketch} & \textbf{Means}  \\ \midrule
\multirow{5}{*}{\textbf{Control$^\downarrow$}} & \textbf{MVControl \cite{li2024controllable}}   & 0.0952        & 0.0160         & 0.0315          & 0.0370          & 0.0449          \\
                                  & \textbf{ControLRM-T \cite{xu2024controlrm}} & 0.0607        & 0.0599         & 0.0359          & 0.0202          & 0.0442          \\
                                  & \textbf{ControLRM-D \cite{xu2024controlrm}} & 0.0600        & 0.0650         & 0.0357          & 0.0197          & 0.0451          \\
                                  & \textbf{CyC3D-T (Ours)} & 0.0342        & 0.0065         & 0.0062          & 0.0159          & 0.0157          \\
                                  & \textbf{CyC3D-D (Ours)} & 0.0347        & 0.0060         & 0.0060          & 0.0156          & \textbf{0.0156} \\ \midrule
\multirow{5}{*}{\textbf{FID$^\downarrow$}}     & \textbf{MVControl \cite{li2024controllable}}   & 311.91        & 256.08         & 255.68          & 288.64          & 278.08          \\
                                  & \textbf{ControLRM-T \cite{xu2024controlrm}} & 253.59        & 262.95         & 269.62          & 256.84          & 260.75          \\
                                  & \textbf{ControLRM-D \cite{xu2024controlrm}} & 165.38        & 174.29         & 174.98          & 169.88          & 171.13          \\
                                  & \textbf{CyC3D-T (Ours)} & 258.16        & 248.84         & 250.87          & 263.32          & 255.30          \\
                                  & \textbf{CyC3D-D (Ours)} & 168.11        & 168.49         & 167.78          & 170.46          & \textbf{168.71} \\ \midrule
\multirow{5}{*}{\textbf{CLIP-I$^\uparrow$}}  & \textbf{MVControl \cite{li2024controllable}}   & 0.762         & 0.841          & 0.851           & 0.811           & 0.816           \\
                                  & \textbf{ControLRM-T \cite{xu2024controlrm}} & 0.855         & 0.842          & 0.835           & 0.852           & 0.846           \\
                                  & \textbf{ControLRM-D \cite{xu2024controlrm}} & 0.854         & 0.826          & 0.820           & 0.850           & 0.838           \\
                                  & \textbf{CyC3D-T (Ours)} & 0.847         & 0.828          & 0.847           & 0.839           & 0.840           \\
                                  & \textbf{CyC3D-D (Ours)} & 0.848         & 0.846          & 0.843           & 0.846           & \textbf{0.846}  \\ \midrule
\multirow{5}{*}{\textbf{CLIP-T$^\uparrow$}}  & \textbf{MVControl \cite{li2024controllable}}   & 0.263         & 0.298          & 0.299           & 0.291           & 0.288           \\
                                  & \textbf{ControLRM-T \cite{xu2024controlrm}} & 0.293         & 0.288          & 0.284           & 0.291           & 0.289           \\
                                  & \textbf{ControLRM-D \cite{xu2024controlrm}} & 0.304         & 0.301          & 0.300           & 0.303           & 0.302           \\
                                  & \textbf{CyC3D-T (Ours)} & 0.296         & 0.296          & 0.295           & 0.292           & 0.295           \\
                                  & \textbf{CyC3D-D (Ours)} & 0.303         & 0.304          & 0.303           & 0.302           & \textbf{0.303}  \\ \bottomrule
\end{tabular}
}
\vspace{-0.3cm}
\caption{Comparison with state-of-the-art controllable 3D generation models under each conditional controls on \textbf{GOBJ$\rightarrow$GSO} benchmark. $\uparrow$ means higher result is better, while $\downarrow$ means lower is better.}
\vspace{-0.3cm}
\label{tab:comparison_each_condition}
\end{table}

In this section, we further provide the quantitative comparison with other state-of-the-art controllable 3D generation methods under different conditional controls in Tab. \ref{tab:comparison_each_condition}.
We compare these methods with controllability metrics and generation quality metrics.
For controllability, we utilize the modified \textbf{MSE} for Edge/Sketch condition, \textbf{R-MSE} for Depth condition, \textbf{DN-CON} for Normal condition.
For generation quality, we report the \textbf{FID}, \textbf{CLIP-I}, \textbf{CLIP-T} under \textbf{All Views} setting.
The table shows the benchmark results on \textbf{GOBJ$\rightarrow$GSO}.
From the table, we can find:
Our proposed CyC3D can effectively improve the controllability on all conditional controls in the table, meantime preserving competitive generation quality compared with other state-of-the-art methods.

\subsection*{Additional Ablation Results}

\begin{figure}[t]
    \centering
    \includegraphics[width=\linewidth]{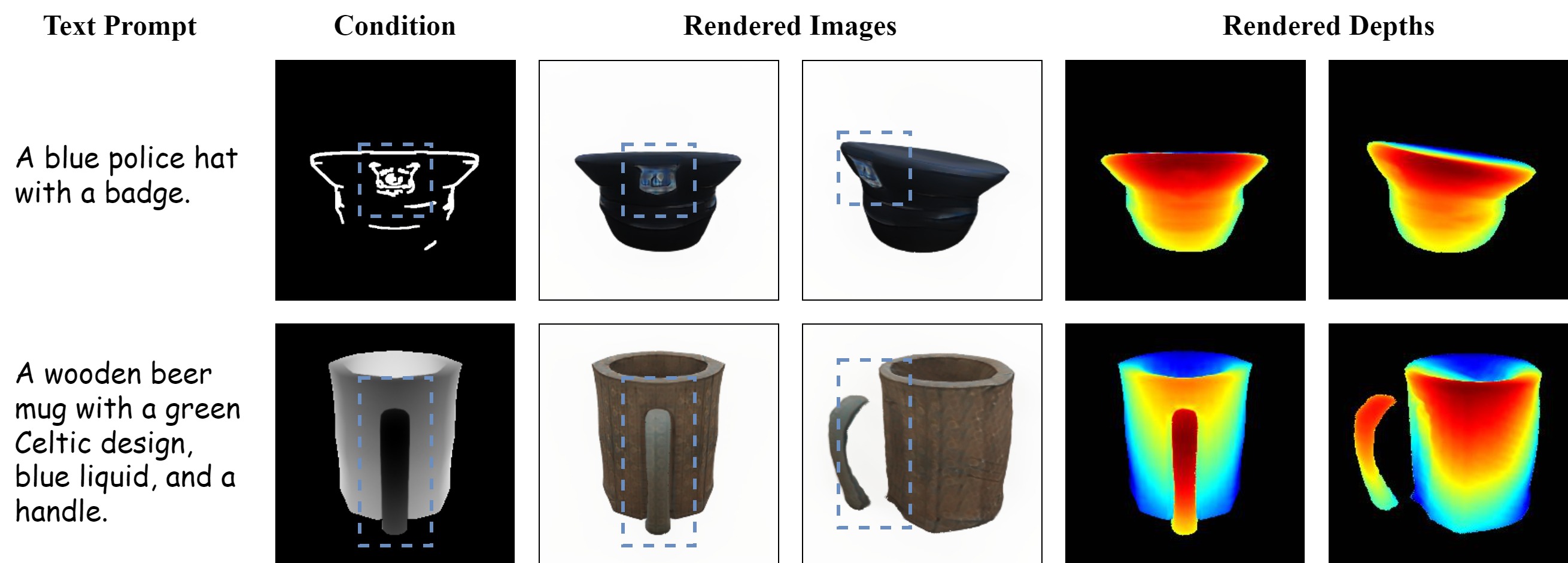}
    \caption{Failure Cases.}
    \label{fig:failure_cases}
\end{figure}

\begin{figure*}[t]
    \centering
    \includegraphics[width=\linewidth]{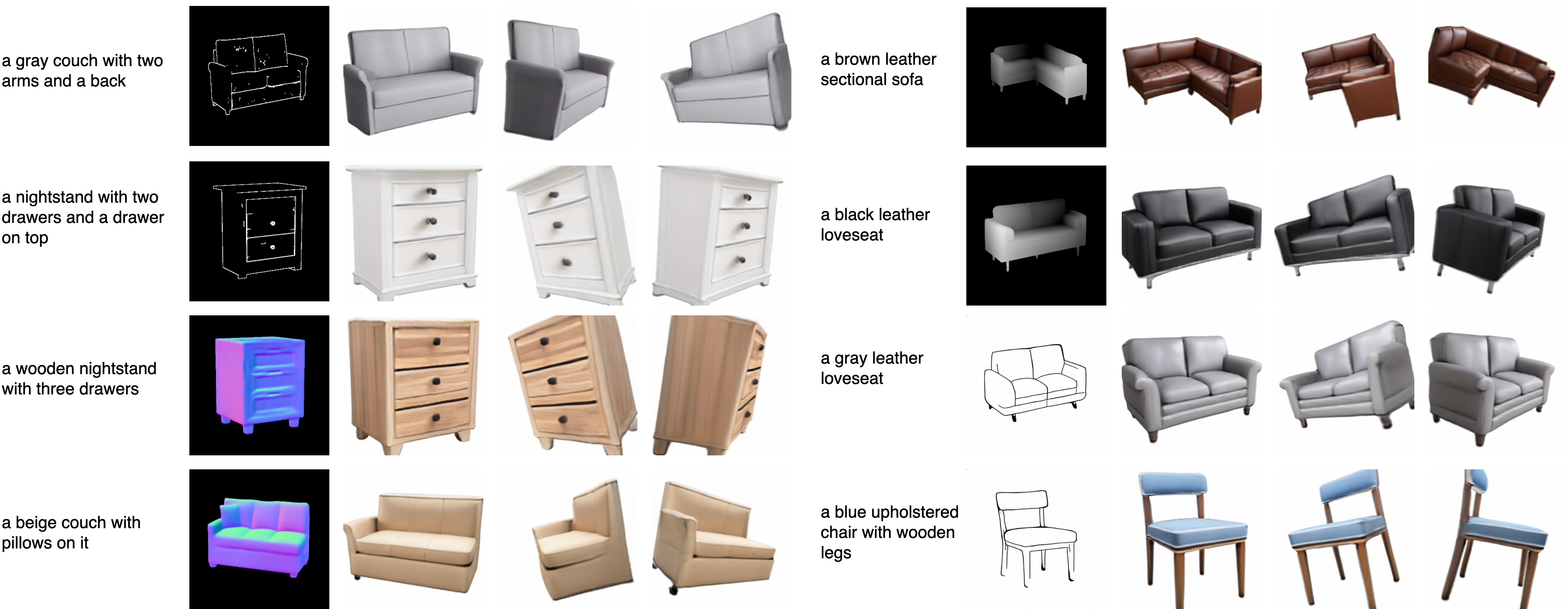}
    \caption{More visualization results.}
    \label{fig:visualization}
\end{figure*}

\noindent\underline{\textbf{Ablation of Different Backbones}}
To find out the performance when different backbones are adopted, we can respectively compare CyC3D-T/-D with ControLRM-T/-D in Tab. \ref{tab:comparison_each_condition}.
Our CyC3D adopts the same network architectures as ControLRM under 2 different network backbones (transformer/diffusion), noted as -T/-D.
From the experimental results, we can find that both CyC3D-T and CyC3D-D achieve better controllability scores compared with ControLRM-T and ControLRM-D.
Furthermore, the generation quality (FID, CLIP-I, CLIP-T) of our CyC3D-T and CyC3D-D is competitive and even better than ControLRM-T and ControLRM-D.
It demonstrates the effectiveness of our proposed method on different backbone networks, including transformer and diffusion backbones.

\begin{table}[t]
\centering
\resizebox{\linewidth}{!}{
\begin{tabular}{c|ccc|ccc}
\toprule
\multirow{2}{*}{\textbf{Loss}} & \multicolumn{3}{c|}{\textbf{Controllability}} & \multicolumn{3}{c}{\textbf{Generation Quality}}  \\
                               & \textbf{PSNR$^\uparrow$} & \textbf{SSIM$^\uparrow$} & \textbf{MSE$^\downarrow$} & \textbf{FID$^\downarrow$} & \textbf{CLIP-I$^\uparrow$} & \textbf{CLIP-T$^\uparrow$} \\ \midrule
\textbf{w/o $L_\text{cond}$}             & 12.87       & 0.841        & 0.0570       & 165.38       & 0.854           & 0.304           \\
\textbf{w $L_\text{cond}$}             & 15.09       & 0.885        & 0.0347       & 168.71       & 0.846           & 0.303          \\ \bottomrule
\end{tabular}
}
\vspace{-0.3cm}
\caption{Ablation experiments of each loss terms under edge conditional control on \textbf{GOBJ$\rightarrow$GSO} benchmark. $\uparrow$ means higher result is better, while $\downarrow$ means lower is better.}
\vspace{-0.3cm}
\label{tab:ablation_edge}
\end{table}

\begin{table}[t]
\centering
\resizebox{\linewidth}{!}{
\begin{tabular}{c|ccc|ccc}
\toprule
\multirow{2}{*}{\textbf{Loss}} & \multicolumn{3}{c|}{\textbf{Controllability}} & \multicolumn{3}{c}{\textbf{Generation Quality}}  \\
                               & \textbf{PSNR$^\uparrow$} & \textbf{SSIM$^\uparrow$} & \textbf{MSE$^\downarrow$} & \textbf{FID$^\downarrow$} & \textbf{CLIP-I$^\uparrow$} & \textbf{CLIP-T$^\uparrow$} \\ \midrule
\textbf{w/o $L_\text{cond}$}             & 17.52         & 0.888         & 0.0187       & 169.88       & 0.850           & 0.303           \\
\textbf{w $L_\text{cond}$}             & 18.38         & 0.890         & 0.0156       & 170.46       & 0.846           & 0.302           \\ \bottomrule
\end{tabular}
}
\vspace{-0.3cm}
\caption{Ablation experiments of each loss terms under sketch conditional control on \textbf{GOBJ$\rightarrow$GSO} benchmark. $\uparrow$ means higher result is better, while $\downarrow$ means lower is better.}
\vspace{-0.3cm}
\label{tab:ablation_sketch}
\end{table}

\begin{table}[t]
\centering
\resizebox{\linewidth}{!}{
\begin{tabular}{c|ccc|ccc}
\toprule
\multirow{2}{*}{\textbf{Loss}} & \multicolumn{3}{c|}{\textbf{Controllability}}     & \multicolumn{3}{c}{\textbf{Generation Quality}}  \\
                               & \textbf{M-MSE$^\downarrow$} & \textbf{Z-MSE$^\downarrow$} & \textbf{R-MSE$^\downarrow$} & \textbf{FID$^\downarrow$} & \textbf{CLIP-I$^\uparrow$} & \textbf{CLIP-T$^\uparrow$} \\ \midrule
\textbf{w/o $L_\text{cond}$}             & 0.0383         & 0.0546         & 0.0650         & 174.29       & 0.826           & 0.301           \\
\textbf{w $L_\text{cond}$}             & 0.0192         & 0.0241         & 0.0321         & 172.10       & 0.832           & 0.302           \\
\textbf{w $L_\text{cond}+L_{\text{cond-d}}$}      & 0.0051         & 0.0041         & 0.0060         & 168.49       & 0.846           & 0.304           \\ \bottomrule
\end{tabular}
}
\vspace{-0.3cm}
\caption{Ablation experiments of each loss terms under depth conditional control on \textbf{GOBJ$\rightarrow$GSO} benchmark. $\uparrow$ means higher result is better, while $\downarrow$ means lower is better.}
\vspace{-0.3cm}
\label{tab:ablation_depth}
\end{table}

\begin{table}[t]
\centering
\resizebox{\linewidth}{!}{
\begin{tabular}{c|cc|ccc}
\toprule
\multirow{2}{*}{\textbf{Loss}} & \multicolumn{2}{c|}{\textbf{Controllability}} & \multicolumn{3}{c}{\textbf{Generation Quality}}  \\
                               & \textbf{NB-MSE$^\downarrow$}       & \textbf{DN-CON$^\downarrow$}      & \textbf{FID$^\downarrow$} & \textbf{CLIP-I$^\uparrow$} & \textbf{CLIP-T$^\uparrow$} \\ \midrule
\textbf{w/o $L_\text{cond}$}             & 0.0034                & 0.0205               & 174.98       & 0.820           & 0.300           \\
\textbf{w $L_\text{cond}$}             & 0.0023                & 0.0187               & 172.18       & 0.831           & 0.302           \\
\textbf{w $L_\text{cond} + L_\text{cond-n}$}      & 0.0011                & 0.0056               & 167.78       & 0.843           & 0.303           \\ \bottomrule
\end{tabular}
}
\vspace{-0.3cm}
\caption{Ablation experiments of each loss terms under normal conditional control on \textbf{GOBJ$\rightarrow$GSO} benchmark. $\uparrow$ means higher result is better, while $\downarrow$ means lower is better.}
\vspace{-0.3cm}
\label{tab:ablation_normal}
\end{table}

\noindent\underline{\textbf{Ablation of Loss Terms under Edge Condition}}
In Tab. \ref{tab:ablation_edge}, we present the ablation results of the loss terms under edge condition control.
Both metrics of controllability and generation quality are reported in the table.
The experiments are conducted on $\textbf{GOBJ$\rightarrow$GSO}$ benchmark.
From the results, we can find that:
(1) The proposed cycle consistency can significantly improve the controllability.
(2) The generation quality might be slightly worse than the baseline by 1.97\% in FID, 0.95\% in CLIP-I, and 0.33\% in CLIP-T.
In summary, our CyC3D can effectively improve the controllability while preserving competitive generation quality under edge conditional control.

\noindent\underline{\textbf{Ablation of Loss Terms under Sketch Condition}}
In Tab. \ref{tab:ablation_edge}, we present the ablation results of the loss terms under sketch condition control.
Both metrics of controllability and generation quality are reported in the table.
The experiments are conducted on $\textbf{GOBJ$\rightarrow$GSO}$ benchmark.
From the results, we can find that:
(1) The proposed cycle consistency can significantly improve the controllability.
(2) The generation quality might be slightly worse than the baseline by 0.34\% in FID, 0.47\% in CLIP-I, and 0.33\% in CLIP-T.
In summary, our CyC3D can effectively improve the controllability while preserving competitive generation quality under sketch conditional control.

\noindent\underline{\textbf{Ablation of Loss Terms under Depth Condition}}
In Tab. \ref{tab:ablation_edge}, we present the ablation results of the loss terms under depth condition control.
Both metrics of controllability and generation quality are reported in the table.
The experiments are conducted on $\textbf{GOBJ$\rightarrow$GSO}$ benchmark.
The findings are summarized as follows:
(1) Siginificant improvements in depth controllability can be observed in each loss term.
(2) The generation quality is also improved slightly compared with the baseline.
In summary, our CyC3D can siginificantly improve the controllability meantime ensuring competitive and even better generation quality under depth conditional control.

\noindent\underline{\textbf{Ablation of Loss Terms under Normal Condition}}
In Tab. \ref{tab:ablation_edge}, we present the ablation results of the loss terms under normal condition control.
Both metrics of controllability and generation quality are reported in the table.
The experiments are conducted on $\textbf{GOBJ$\rightarrow$GSO}$ benchmark.
The findings are summarized as follows:
(1) Siginificant improvements in depth-normal consistency can be observed in each loss term.
(2) The generation quality is also improved slightly compared with the baseline.
In summary, our CyC3D can siginificantly improve the controllability meantime ensuring competitive and even better generation quality under normal conditional control.

\section*{Discussion}
\label{suppl:failure}

\noindent\underline{\textbf{Improvement on Depth/Normal vs Edge/Sketch}}
From the quantitative and qualitative results, we can observe that: the improvement of our proposed CyC3D on depth/normal condition control is much higher than the ones on edge/sketch conditional control.
We summarize two important issues as follows:

(1) \textbf{Alignment of correct 3D condition maps with the ground truth 3D object}.
In existing methods (like MVControl \cite{li2024controllable}), the depth and normal condition maps are usually extracted from images with pre-trained models.
However, these models tend to predict over-smoothed depth or normal maps.
Furthermore, these generated depth or normal maps have different scales with the real 3D object, leading to incorrect 3D shape in these generated condition maps.
Since these incorrect depth or normal maps are set as input, the training phase will enforce the generated 3D shapes to align with these incorrect 3D conditions.
Consequently, this problem may significantly degrade the final performance of controllable 3D generation models.
Different from existing methods, our CyC3D can effectively align the 3D representation and the input 3D conditons like depth and normal.
It can not only mitigate the difference in scale, but also avoid the discordance between the 3D condtion and the 3D object.
Consequently, our CyC3D can improve the controllability significantly meantime providing competitive and even better generation quality.

(2) \textbf{2D condition maps like edge/sketch may have ambiguities when mapping from 2D to 3D.}
During the preparation edge or sketch condition maps in the dataset, the extracted edge or sketch conditions are usually noisy and sensitive to hyper-parameters or disturbances in images like light conditions.
This leads to an unexpected fact that the noisy condition maps might contain ambiguous information.
For example, some reflection of lights might leads to fake edges or sketchs in the condition maps, some noise in the images might create dots-like edges or sketchs in the condition maps.
In consequence, the natural disturbance in edge/sketch maps might involve extra noise during training, thus leading to degradation of generation quality slightly.
But the controllability is not affected by this issue, because it only concentrates on the cycle consistency between input condition and generated results.
Hence, the controllability is improved significantly normally.

\noindent\underline{\textbf{Difference with Cycle3D}}
Note that Cycle3D \cite{tang2025cycle3d} also applies the idea of cycle consistency in image-to-3D generation tasks, achieving superior performance.
The differences between our work and Cycle3D are summarized as follows:
(1) Cycle3D builds the cycle consistency on the pipeline of: diffusion model $\rightarrow$ reconstruction model $\rightarrow$ diffusion model --> ... It is an inference pipeline for generating corresponding 3D contents from given image inputs. However, our work is totally different. Our CyC3D applies the cycle consistency constraint of conditional inputs during the training phase. 
Whereas our CyC3D directly generates the results with one feed-forward model during inference, rather than propagating through multiple rounds of diffusion and reconstruction models in Cycle3D.
(2) Our CyC3D handles a different task compared with Cycle3D. We aim to handle the controllable 3D generation task with conditional inputs like edge, depth, normal or sketch maps. Whereas Cycle3D only handles the image-to-3D generation problem.

\noindent\underline{\textbf{Failure Cases}}
In Fig. \ref{fig:failure_cases}, we show some failure cases of our model.
The first column is the text prompt, and the second column is the input condition map.
The remaining columns visualize the rendered images and depths at different viewpoints.
(1) As the the first row shows, due to the strong regularization on the controllability of our CyC3D, the blurry and noisy edge condition in the blue dotted box might make the model confusing to generate a reasonable result.
(2) As the second row shows, the generated result seems to be reasonable on the same reference view, but fails at other viewpoints.
The blue dotted box highlights the failures of the generated 3D object.

\noindent\underline{\textbf{Limitation}}
Based on aforementioned analysis on failure cases in Fig. \ref{fig:failure_cases}, we can summarize the limitation of our proposed as follows:
(1) Our model requires accurate conditional control, otherwise the blurry conditions might lead to unsatisfying generated results.
Because the strong controllability ability of CyC3D might reversely be confused by inaccurate or blurry condition maps.
A future direction is how to mitigate these disturbances in condition maps.
(2) The bottleneck of our model depends on the bottleneck of the pre-trained backbone (i.e. OpenLRM).
Since we directly inherits the parameters of triplane transformers in OpenLRM-base, the bottleneck of the backbone model (OpenLRM-base) limits our model as well.
The performance of our model depends on the performance of the backbone model.
Replacing the backbone with a large and powerful model can increase the performance significantly.

\end{document}